\documentclass[10pt,twocolumn,letterpaper]{article}
\usepackage{cvpr}
\usepackage{times}
\usepackage{epsfig}
\usepackage{graphicx}
\usepackage{amsmath}
\usepackage{amssymb}
\usepackage{subcaption}
\usepackage{multirow}
\usepackage{enumitem}
\usepackage{booktabs}
\usepackage{pifont}
\newcommand{\cmark}{\textcolor{green}{\ding{51}}}%
\newcommand{\xmark}{\textcolor{red}{\ding{55}}}
\usepackage[toc,page]{appendix}

\usepackage[pagebackref=true,breaklinks=true,letterpaper=true,colorlinks,bookmarks=false]{hyperref}
\cvprfinalcopy 

\def\cvprPaperID{2040} 

\newcommand{\new}[1]{#1}

\begin{document}

\title{DVQA: Understanding Data Visualizations via Question Answering}

\author{Kushal~Kafle$^{1,}$\thanks{A portion of this research was done while Kushal Kafle was an intern at Adobe Research.} \qquad Brian~Price$^2$
  \qquad Scott~Cohen$^2$ \qquad Christopher~Kanan$^1$\\
  \hfill$^1$Rochester Institute of Technology\hfill
  $^2$Adobe Research\hfill\mbox{ }\\
{\tt\small $^1$\{kk6055, kanan\}@rit.edu \qquad  $^2$\{bprice, scohen\}@adobe.com}}

\maketitle

\begin{abstract}
Bar charts are an effective way to convey numeric information, but today's algorithms cannot parse them. Existing methods fail when faced with even minor variations in appearance. Here, we present DVQA, a dataset that tests many aspects of bar chart understanding in a question answering framework. Unlike visual question answering (VQA), DVQA requires processing words and answers that are unique to a particular bar chart. State-of-the-art VQA algorithms perform poorly on DVQA, and we propose two strong baselines that perform considerably better. Our work will enable algorithms to automatically extract numeric and semantic information from vast quantities of bar charts found in scientific publications, Internet articles, business reports, and many other areas. 
\end{abstract}

\section{Introduction}\label{sec:intro}

\begin{figure}[t]

 \captionsetup[subfigure]{labelformat=empty,font=footnotesize}    
        \begin{subfigure}[t]{0.49\textwidth}
		\includegraphics[width=\textwidth]{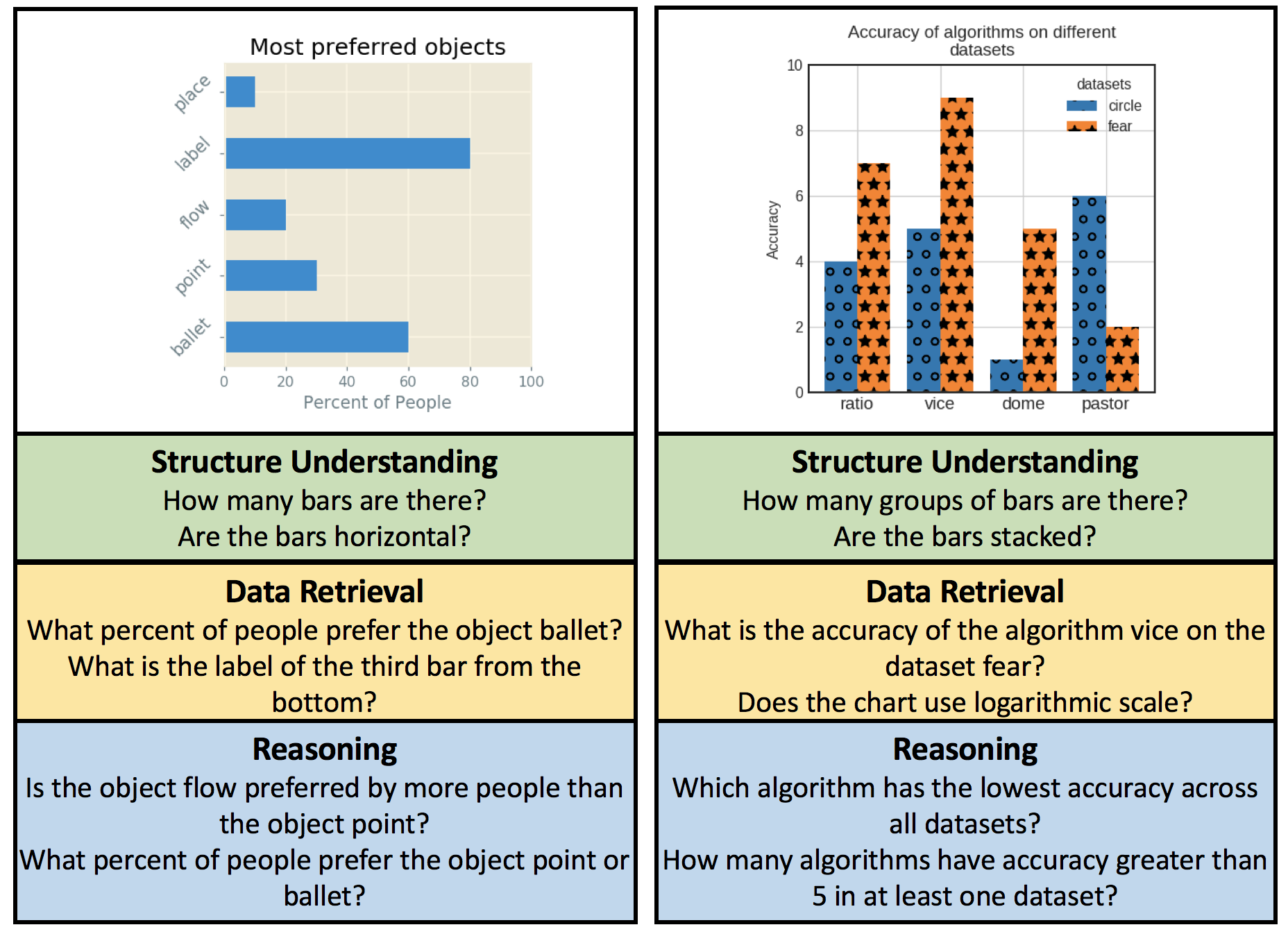}
    \end{subfigure}
   \caption{\small DVQA involves answering questions about diagrams. We present a dataset for DVQA with bar charts that exhibit enormous variety in appearance and style. We show that VQA systems cannot answer many DVQA questions and we describe more effective algorithms. \label{fig:overview}}
\end{figure}

Data visualizations, \eg, bar charts, pie charts, and plots, contain large amounts of information in a concise format. These visualizations are specifically designed to communicate data to people, and are not designed to be machine interpretable. Nevertheless, making algorithms capable to make inferences from data visualizations has enormous practical applications. Here, we study systems capable of answering open-ended questions about bar charts, which we refer to as data visualization question answering (DVQA). DVQA would enable vast repositories of charts within scientific documents, web-pages, and business reports to be queried automatically. Example DVQA images and questions grouped by the different tasks are shown in Fig.~\ref{fig:overview}.

\new{Besides practical benefits, DVQA can also serve as a challenging proxy task for generalized pattern matching, attention, and multi-step reasoning systems. Answering a question about a chart requires multi-step attention, memory, measurement, and reasoning that poses significant challenges to the existing systems. For example, to answer the question \textit{`What is the accuracy of algorithm} vice \textit{on the dataset} fear\textit{?'} in Fig.~\ref{fig:overview} can require finding the appropriate color and hatching that represents the dataset \textit{fear}, finding the group of bars that represent the algorithm \textit{vice}, measuring the height of the bar based on the y-axis, and if necessary interpolating between two neighboring values.}

\begin{figure}[t]
    \captionsetup[subfigure]{labelformat=empty}    
        \begin{subfigure}[t]{0.23\textwidth}
		\includegraphics[width=\textwidth, height=1.5in]{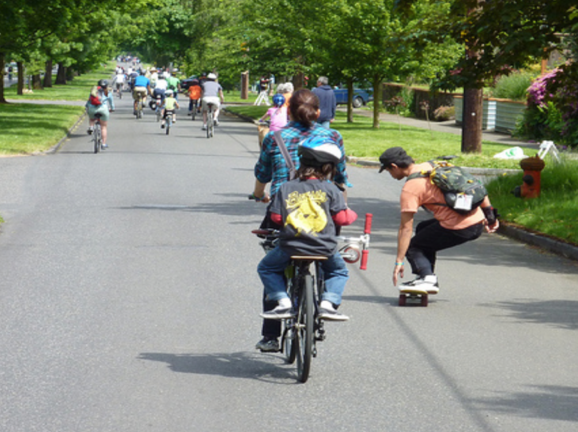}
    \end{subfigure}
    \hfill
    \begin{subfigure}[t]{0.23\textwidth}
	 	\includegraphics[width=\textwidth, height=1.5in]{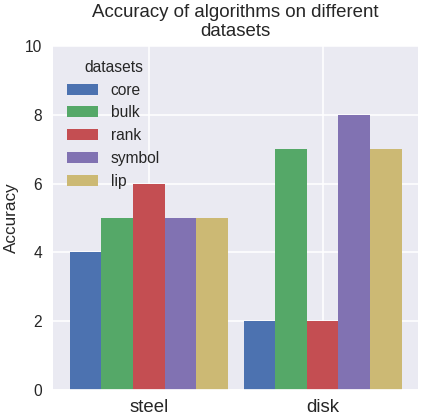}
	\end{subfigure}
    \caption{\small
Natural images vs. bar charts. \textbf{Left}: Small changes in an image typically have little impact on a question in VQA. \textbf{Right:} Bar charts convey information using a sparse, but precise, set of visual elements. Even small changes can completely alter the information in the chart.}
    \label{fig:diff}
    \end{figure}

DVQA is related to visual question answering (VQA)~\cite{malinowski2014multi,antol2015vqa}, which deals with answering open-ended questions about images. VQA is usually treated as a classification problem, in which answers are categories that are inferred using features from image-question pairs. DVQA poses three major challenges that are overlooked by  existing VQA datasets with natural images. First, VQA systems typically assume two fixed vocabulary dictionaries: one for encoding words in questions and one for producing answers. In DVQA, assuming a fixed vocabulary makes it impossible to properly process many questions or to generate answers unique to a bar chart, which are often labeled with proper nouns, abbreviations, or concatenations (\eg, `Jan-Apr'). Our models demonstrate two ways for handling out-of-vocabulary (OOV) words. \new{Second, the language utilized in VQA systems represent fixed semantic concepts that are immutable over images, \eg, phrases such as `A large shiny red cube' used in CLEVR~\cite{johnson2016clevr} represent a fixed concept; once the word `red' is associated with the underlying semantic concept, it is immutable. By contrast, the words utilized in labels and legends in DVQA can often be arbitrary and could refer to bars of different position, size, texture, and color.} Third, VQA's natural images exhibit regularities that are not present in DVQA, \eg to infer the answer to `What is the weather like?' for the image in Fig.~\ref{fig:diff}, an agent could use color tones and overall brightness to infer `sunny.' Changing the color of the fire hydrant only changes the local information that impacts questions about the fire hydrant's properties. However, in bar charts, a small change, \eg, shuffling the colors of the legend in Fig.~\ref{fig:diff}, completely alters the chart's information. This makes DVQA an especially challenging problem.

\textbf{This paper makes three major contributions:}
\begin{enumerate}[noitemsep,nolistsep]
\item We describe the DVQA dataset, which contains over 3 million image-question pairs about bar charts. It tests three forms of diagram understanding:  a) structure understanding; b) data retrieval; and c) reasoning. The DVQA dataset will be publicly released.
\item We show that both baseline and state-of-the-art VQA algorithms are incapable of answering many of the questions in DVQA. Moreover, existing classification based systems based on a static and predefined vocabulary are incapable of answering questions with unique answers that are not encountered during training.
\item We describe two DVQA systems capable of handling words that are unique to a particular image. One is an end-to-end neural network that can read answers from the bar chart. The second is a model that encodes a bar chart's text using a dynamic local dictionary.
\end{enumerate}

\section{Related Work}
\subsection{Automatically Parsing Bar Charts}
Extracting data from bar charts using computer vision has been extensively studied~\cite{al2015automatic,elzer2011automated,kallimani2013extraction,poco2017reverse,savva2011revision}. Some focus on extracting the visual elements from the bar charts~\cite{poco2017reverse}, while others focus on extracting the data from each bar directly~\cite{savva2011revision,kallimani2013extraction}. Most of these approaches use fixed heuristics and make strong simplifying assumptions, \eg, \cite{savva2011revision} made  several simplifying assumptions about bar chart appearance (bars are solidly shaded without textures or gradients, no stacked bars, etc.). Moreover, they only tested their data extraction procedure on a total of 41 bar charts.

Our DVQA dataset has variations in bar chart appearance that go far beyond the capabilities of any of the aforementioned works. Moreover, DVQA requires more than just data extraction. Correctly answering DVQA questions requires basic language understanding, attention, concept of working short-term memory and reasoning.

\subsection{VQA with Natural Images}
Over the past three years, multiple VQA datasets containing natural images have been publicly released~\cite{malinowski2014multi,antol2015vqa,ren2015image,krishnavisualgenome,kafle2017analysis}. The most popular dataset is The VQA Dataset~\cite{balanced_vqa_v2,antol2015vqa}. It is much larger and more varied than earlier VQA datasets, such as COCO-QA~\cite{ren2015image} and DAQUAR~\cite{malinowski2014multi}. However, the first version of the dataset, VQA 1.0, suffered from extreme language bias, resulting in many questions not requiring the image to correctly answer them~\cite{balanced_vqa_v2}. In the second version, VQA 2.0, this bias was greatly reduced; however, VQA 2.0 still suffers from heavily skewed distribution in the \emph{kinds} of questions present in the dataset~\cite{kafle2017analysis}.

Numerous VQA algorithms have been proposed, ranging from Bayesian approaches~\cite{kafle2016,malinowski2014multi}, methods using spatial attention~\cite{Yang2016,xu2015show,LuYBP16,noh2016training}, compositional approaches~\cite{AndreasRDK15,andreas2016learning}, and bilinear pooling schemes~\cite{kim2016hadamard,FukuiPYRDR16}. Almost all VQA algorithms pose it as a classification problem in which each class is synonymous with a particular answer. For more extensive reviews see \cite{kafle2016review} and \cite{wu2016visual}.

While there are significant similarities between VQA and DVQA, one critical difference is that many DVQA questions require directly reading text from a chart to correctly answer them. This demands being able to handle words that are unique to a particular chart, which is a capability that is not needed by algorithms operating on existing VQA datasets with natural images. 

\subsection{Reasoning, Synthetic Scenes, and Diagrams}

While VQA is primarily studied using natural images, several datasets have been proposed that use synthetic scenes or diagrams to test reasoning and understanding~\cite{johnson2016clevr,kembhavi2016diagram,Kembhavi2017tqa}. The CLEVR~\cite{johnson2016clevr} dataset has complex reasoning questions about synthetically created scenes, and systems that perform well on popular VQA datasets perform poorly on CLEVR. The TQA~\cite{Kembhavi2017tqa} and AI2D~\cite{kembhavi2016diagram} datasets both involve answering science questions about text and images. Both datasets are relatively small, \eg, AI2D only contains 15,000 questions. These datasets require more than simple pattern matching and memorization. Similar to our work, their creators showed that state-of-the-art VQA systems for natural image datasets performed poorly on their datasets. However, there are key differences between these datasets and DVQA. First, none of these datasets contain questions specific to bar charts. Second, their datasets use multiple-choice schemes that reduce the problem to a ranking problem, rather than the challenges posed by having to generate open-ended answers. Studies have shown that multiple-choice schemes have biases that models will learn to exploit~\cite{jabri2016revisiting}. In contrast, we treat DVQA as an open-ended question answering task.

\new{Concurrent to our work, FigureQA~\cite{figureqa} also explores question answering for charts, however, with following major limitations compared to our DVQA dataset: 1) it contains only yes/no type questions; 2) it does not contain questions that require numeric values as answers; 3) it has fixed labels for bars across different figures (\eg, a red bar is always labeled 'red'); and 4) it avoids the OOV problem.}

\section{DVQA: The Dataset}

\begin{figure*}[!t]
\footnotesize
	\centering
    \captionsetup[subfigure]{labelformat=empty}    
        \begin{subfigure}[t]{0.19\textwidth}
		\includegraphics[width=\textwidth]{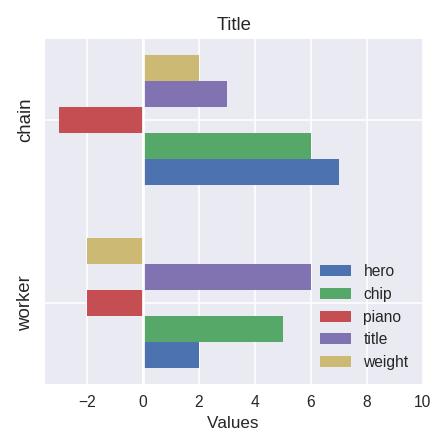}
    \hfill
    \end{subfigure}
           \begin{subfigure}[t]{0.19\textwidth}
		\includegraphics[width=\textwidth]{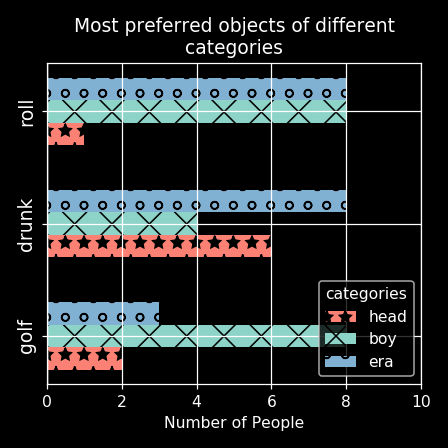}
    \hfill
    \end{subfigure}
           \begin{subfigure}[t]{0.19\textwidth}
		\includegraphics[width=\textwidth]{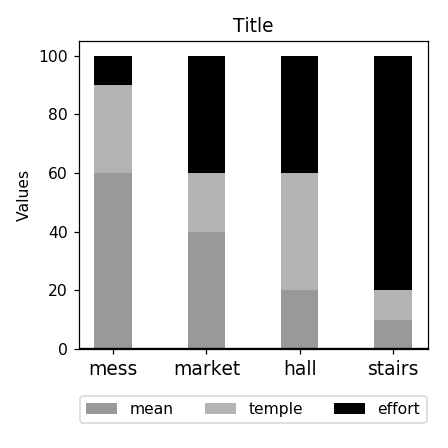}
    \hfill
    \end{subfigure}
        \begin{subfigure}[t]{0.19\textwidth}
		\includegraphics[width=\textwidth]{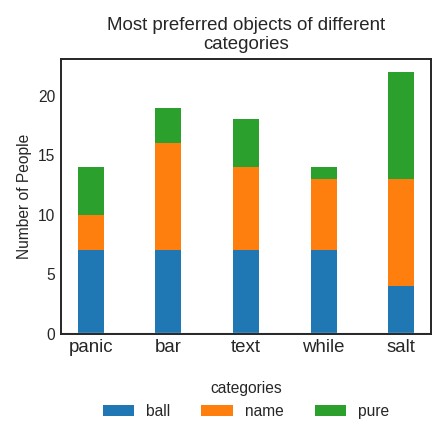}
    \hfill
    \end{subfigure}
           \begin{subfigure}[t]{0.19\textwidth}
		\includegraphics[width=\textwidth]{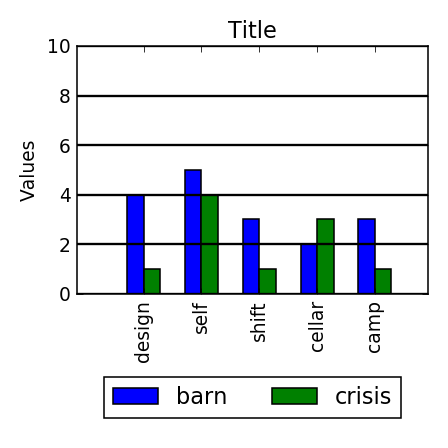}
    \hfill
    \end{subfigure}
   
   \vspace{-15pt}
    \caption{\small Example bar chart images from DVQA. DVQA contains significant variation in appearance and style.}
    \label{fig:examples}
    \end{figure*}

DVQA is a challenging synthetic dataset that tests multiple aspects of bar chart understanding that cause state-of-the-art methods for VQA to fail, which we demonstrate in experiments. Synthetically generating DVQA gave us precise control over the positions and appearances of the visual elements. It also gave us access to meta-data about these components, which would not be available with real data. This meta-data contains all information within the chart, including the precise position of each drawing element, the underlying data used to create the chart and location of all text-elements. This data can be used as an additional source of supervision or to ensure that an algorithm is ``attending'' to relevant regions. As shown in Fig.~\ref{fig:examples}, the DVQA dataset contains a large variety of typically available styles of bar chart. The questions in the dataset require the ability to reason about the information within a bar chart (see Fig.~\ref{fig:overview}).  DVQA contains 3,487,194 total question answer pairs for 300,000 images divided into three major question types. Tables~\ref{tbl:stat-qtypes} and \ref{tbl:stats-overall} show  statistics about the DVQA dataset. Additional statistics are given in the supplemental materials.

\subsection{Appearance, Data, and Question Types}
DVQA consists of bar charts with question-answer pairs that are generated by selecting a visual style for a chart, choosing data for a chart, and then generating questions for that chart. Here, we briefly explain how this was done. Additional details are provided in the supplemental materials.

\textbf{Visual Styles: }\new{We use python's popular drawing tool, Matplotlib to generate our charts since it offers unparalleled programmatic control over each of the element drawn. As shown in Fig.~\ref{fig:examples}, DVQA's bar charts contain a wide variability in both appearance and style that can capture the common styles found in scientific documents and the Internet. Some of these variations include the difference in the number of bars and groups; presence or absence of grid lines; difference in color, width, spacing, orientation, and texture of the bars; and difference in the orientation and the location of labels and legends.}

To label individual bars and legend entries, we select the 1000 most frequent nouns in the Brown Corpus using NLTK's part-of-speech tagging for our training set and our `easy' test set \textbf{\textit{Test-Familiar}}. To measure a system's ability to scale to unknown answers, we also created a more difficult test set \textbf{\textit{Test-Novel}}, in which we use 500 new words that are not seen during training.

\textbf{Underlying Data:}
\new{DVQA has three bar chart data types: linear, percentage, and exponential. For each of these data value types, the bars can take any of the 10 randomly chosen values in the range 1 -- 10 for linear data, 10 -- 100 for percentage, and 1 - $10^{10}$ for exponential data type. A small percentage of bars are allowed to have a value of zero which appears as a missing bar in the chart.}

\textbf{Question Types:}
DVQA contains three types of questions: 1) structure understanding, 2) data retrieval, and 3) reasoning. \new{To generate these questions, we use fixed templates. Based on the context of the chart reflected through its title and labels, the questions will vary along the template. Below, we will show a random assortment of these questions with further details presented in the supplementary materials.

\textbf{Structure Understanding.} Structure understanding  questions test a system's ability to understand the overall structure of a bar chart. These questions include: 
\begin{enumerate}[noitemsep, wide, labelwidth=!, labelindent=0pt]
\small
\item How many bars are there?
\item How many groups/stacks of bars are there?
\item How many bars are there per group?
\item Does the chart contain any negative values?
\item Are the bars horizontal?
\item Does the chart contain stacked bars?
\item Is each bar a single solid color without patterns?
\end{enumerate}

\textbf{Data Retrieval.} Data retrieval questions test a system's ability to retrieve information from a bar chart by parsing the chart into its individual components. These questions often require paying attention to specific region of the chart. These questions include:
\begin{enumerate}[noitemsep, wide, labelwidth=!, labelindent=0pt]
\small
\item Are the values in the chart presented in a logarithmic scale?
\item Are the values in the chart presented in a percentage scale?
\item What percentage of people prefer the object \textbf{O}?
\item What is the label of the third bar from the left?
\item What is the label of the first group of bars from the left?
\item What is the label of the second bar from the left in each group?
\item What element does the \textbf{C} color represent?
\item How many units of the item \textbf{I} were sold in the store \textbf{S}?
\end{enumerate}

\textbf{Reasoning.} Reasoning questions test a model's ability to collect information from multiple components of a bar chart  and perform operations on them. These include:

\begin{enumerate}[noitemsep, wide, labelwidth=!, labelindent=0pt]
\small
\item Which algorithm has the highest accuracy?
\item How many items sold more than $N$ units?
\item What is the difference between the largest and the smallest value in the chart?
\item How many algorithms have accuracies higher than \textbf{N}?
\item What is the sum of the values of \textbf{L1} and \textbf{L2}?
\item Did the item \textbf{I1} sold less units than \textbf{I2}?
\item How many groups of bars contain at least one bar with value greater than N?
\item Which item sold the most units in any store?
\item Which item sold the least number of units summed across all the stores?
\item Is the accuracy of the algorithm \textbf{A1} in the dataset \textbf{D1} larger than the accuracy of the algorithm \textbf{A2} in the dataset \textbf{D2}?
\end{enumerate}
}
\subsection{Post-processing to Minimize Bias}

Several studies in VQA have shown that bias in datasets can impair performance evaluation and give inflated scores to systems that simply exploit statistical patterns~\cite{jabri2016revisiting,kafle2017analysis,vqacp}. In DVQA, we have taken several measures to combat such biases. 
To ensure that there is no correlation between styles, colors, and labels, we randomize the generation of charts. Some questions can have strong priors, \eg, the question `Does the chart contain stacked bar?' has a high probability of the correct answer being `no' because these stacked charts are uncommon. To compensate for this, we randomly remove these questions until yes/no answers are balanced for each question type where yes/no is an answer. A similar scheme was used to balance other structure understanding questions as well as the first two data retrieval questions.

\begin{table}[t]
\centering
\footnotesize
\caption{Dataset statistics for different DVQA splits for different question types.}
\vspace{-6pt}
\label{tbl:stat-qtypes}
\begin{tabular}{@{}lrrr@{}}
\toprule
                                & \textbf{\begin{tabular}[c]{@{}l@{}}Total\\ Questions\end{tabular}} & \textbf{\begin{tabular}[c]{@{}l@{}}Unique\\ Answers\end{tabular}} \\ \midrule
Structure          & 471,108                                                           & 10                                                                \\              
Data             & 1,113,704                                                            & 1,538                                                               \\
                                                          
Reasoning          & 1,613,974                                                           & 1,576   \\                                                                                                                     \midrule
\textbf{Grand Total}   & \textbf{3,487,194}                                                 & \textbf{1,576}                                                      \\ \bottomrule
\end{tabular}
\end{table}

\begin{table}[t]
\centering
\footnotesize
\caption{DVQA dataset statistics for different splits.}
\vspace{-6pt}
\label{tbl:stats-overall}
\begin{tabular}{lrrrr}
\toprule
\multicolumn{1}{l}{} & \textbf{Images} & \textbf{Questions} & \textbf{Unique Answers} \\ \midrule
Train                &    200,000    &    2,325,316       &    1,076            \\
Test-Familiar            &   50,000    &     580,557      &      1,075          \\
Test-Novel            &     50,000   &      581,321      &        577        \\ \midrule
\textbf{Grand Total}                &    \textbf{300,000}    &     \textbf{3,487,194}      &  \textbf{1,576} \\ \bottomrule          
\end{tabular}
\end{table}

\section{DVQA Algorithms \& Models}

In this section, we describe two novel deep neural network algorithms along with five baselines. Our proposed algorithms are able to read text from bar charts, giving them the ability to answer questions with chart-specific answers or requiring chart-specific information. 

All of the models that process images use the ImageNet pre-trained ResNet-152~\cite{resnet} CNN with $448 \times 448$ images resulting in a $14 \times 14 \times 2048$ feature tensor, unless otherwise noted. All models that process questions use a 1024 unit single layer LSTM to encode questions, where each word in the question is embedded into a dense 300 dimensional representation. Training details are given in Sec.~\ref{sec:training-details}.

\subsection{Baseline Models}
\label{sec:baseline}

We evaluate five baseline models for DVQA: 
\begin{enumerate}[noitemsep,nolistsep]

\item \textbf{YES}: This model answers `YES' for all questions, which is the most common answer in DVQA by a small margin over `NO'.

\item \textbf{IMG}: A question-blind model. Images are encoded using Resnet using the output of its final convolutional layer after pooling, and then the answer is predicted from them by an MLP with one hidden-layer that has 1,024 units and a softmax output layer.

\item \textbf{QUES}: An image-blind model. It uses the LSTM encoder to embed the question, and then the answer is predicted by an MLP with one hidden-layer that has 1,024 units and a softmax output layer. 

\item \textbf{IMG+QUES}: This is a combination of the QUES and IMG models. It concatenates the LSTM and CNN embeddings, and then feeds them to an MLP with one 1024-unit hidden layer and a softmax output layer. 

\item \textbf{SAN-VQA}: The Stacked Attention Network (SAN)~\cite{Yang2016} for VQA. We use our own implementation of SAN as described by \cite{kazemi2017show}, where it was shown that upgrading the original SAN's image features and a couple small changes produces state-of-the-art results on VQA 1.0 and 2.0. SAN operates on the last CNN convolutional feature maps, where it processes this map attentively using the question embedding from our LSTM-based scheme.

\end{enumerate}
\begin{figure*}[t]
\centering
	\includegraphics[width=0.9\textwidth]{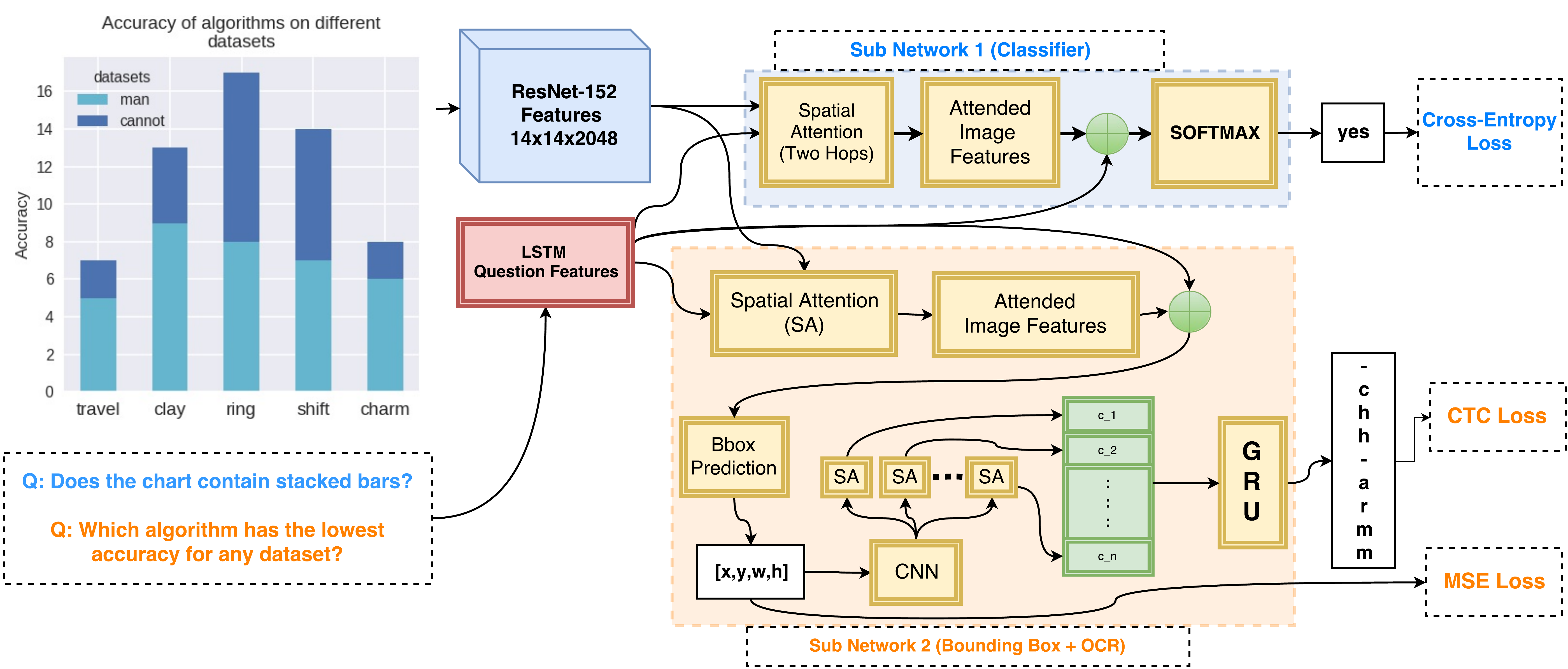}
    \caption{\small \new{Overview of our Multi-Output Model (MOM) for DVQA. MOM uses two sub-networks: 1) classification sub-network that is responsible for \textit{generic answers}, and 2) OCR sub-network that is responsible for \textit{chart-specific answers}.}}
\label{fig:mom}
\end{figure*}

\subsection{Multi-Output Model (MOM)}

Our Multi-Output Model (MOM) for DVQA uses a dual-network architecture, where one of its sub-networks is able to generate chart-specific answers. MOM's classification sub-network is responsible for \textit{generic} answers. MOM's optical character recognition (OCR) sub-network is responsible for \textit{chart-specific} answers that must be read from the bar chart. The classification sub-network is identical to the SAN-VQA algorithm described earlier in Sec.~\ref{sec:baseline}. An overview is given in Fig.~\ref{fig:mom}.

MOM's OCR sub-network tries to predict the bounding box  containing the correct label and then applies a character-level decoder to that region. The bounding box predictor is trained as a regression task using a mean-squared-error (MSE) loss. An image patch is extracted from this region, which is resized to $128\times128$, and then a small 3-layer CNN is applied to it. Since the orientation of the text in the box will vary, we employ an $N$-step spatial attention mechanism to encode the relevant features for each of the $N$ possible characters in the image patch, where $N$ is the largest possible character-sequence ($N=8$ in our experiments). These $N$ features are encoded using a bi-directional gated recurrent unit (GRU) to capture the character level correlations found in naturally occurring words. The GRU encoding is followed by a classification layer that predicts the character sequence, which is trained using connectionist temporal classification (CTC) loss~\cite{graves2006-ctc}.

MOM must determine whether to use the classification sub-network (\ie SAN-VQA) or the OCR sub-network to answer a question. To determine this, we train a separate binary classifier that determines which of the outputs to trust. This classifier takes the LSTM question features as input to predict whether the answer is generic or chart-specific. For our DVQA dataset this classifier is able to predict the correct branch with perfect accuracy on the test data.

\subsection{SANDY: SAN with DYnamic Encoding Model}

MOM handles chart-specific answers by having a sub-network capable of generating unique strings; however, it has no explicit ability to visually read bar chart text and its LSTM question encoding cannot handle chart-specific words. To explore overcoming these limitations, we modified SAN to create SANDY, SAN with DYnamic encoding model. SANDY uses a dynamic encoding model (DEM) that explicitly encodes chart-specific words in the question, and can directly generate chart-specific answers.  The DEM is a  dynamic local dictionary for chart-specific words. This dictionary is used for encoding words as well as answers. 

To create a local word dictionary, DEM assumes it has access to an OCR system that gives it the positions and strings for all text-areas in a bar chart. Given this collection of  boxes, DEM assigns each box a unique numeric index. It assigns an index of $0$ to the box in the lower-left corner of the image. Then, it assigns the box with the position closest to the first box with an index of $1$. The box closest to $1$ that is not yet assigned an index is then assigned the index of $2$, and so on until all boxes in the image are assigned an index. In our implementation, we assume that we have a perfect (oracle) OCR system for input, and we use the dataset's annotations for this purpose. No chart in the training data had more than 30 text labels, so we set the local dictionary to have at most $M=30$ elements.

The local dictionary augments the $N$ element global dictionary. This enables DEM to create ($M+N$)-word dictionary that are used to encode each word in a question. The local dictionary is also used to augment the $L$ element global answer dictionary. This is done by adding $M$ extra classes to the classifier representing the dynamic words. If these classes are predicted, then the output string is assigned using the local dictionary's appropriate index.

We test two versions of SANDY. The oracle version directly uses annotations from the DVQA dataset to build a DEM. The OCR version uses the output of the open-source Tesseract OCR. Tesseract's output is pre-processed in three ways: 1) we only use words with alphabetical characters in them, 2) we filter word detections with confidence less than 50\%, and 3) we filter single-character word detections.

\subsection{Training the Models}
\label{sec:training-details}

All of the classification based systems, except SANDY and the OCR branch of MOM, use a global answer dictionary from training set containing 1076 words, so they each have 1076 output units. MOM's OCR branch contains 27 output units; 1 for each alphabet and and 1 reserved for \texttt{blank} character. Similarly, SANDY's output layer contains 107 units, with the indices 31 through 107 are reserved for common answers and indices 0 through 30 are reserved for the local dictionary.

For a fair comparison, we use the same training hyperparameters for all the models and closely follow the architecture for SAN models from ~\cite{kazemi2017show} wherever possible. SAN portion for all the models are trained using early stopping and regularized using dropout of 0.5 on inputs to all convolutional, fully-connected and LSTM units. All models use Adam~\cite{kingma2014adam} optimizer with an initial learning rate of 0.001.

\begin{table*}[t]
\centering
\footnotesize
\caption{Overall results for models trained and tested on the DVQA dataset. Values are \% of questions answered correctly.}
\vspace{-6pt}
\label{tbl:ques-type}
\begin{tabular}{@{}lrrrr|rrrr@{}}
\toprule
                     & \multicolumn{4}{c|}{\textbf{Test-Familiar}}       & \multicolumn{4}{c}{\textbf{Test-Novel}}       \\ \midrule
\multicolumn{1}{c}{} & Structure & Data & Reasoning & \textbf{Overall} & Structure & Data & Reasoning & \textbf{Overall} \\ \midrule
YES             & 41.14     & 7.45 	& 8.31      & 11.78          & 41.01     & 7.52 	& 8.23      & 11.77          \\
IMG             & 60.09     & 9.07  & 8.27      & 14.83          & 59.83     & 9.11 	& 8.37     	& 14.90            \\
QUES            & 44.03     & 9.82 	& 25.87     & 21.06          & 43.90    & 9.80	& 25.76     	& 21.00          \\
IMG+QUES        & 90.38     & 15.74 & 31.95     & 32.01          & 90.06     & 15.85   & 31.84     & 32.01            \\
SAN-VQA    		& 94.71    	& 18.78	& 37.29    	& 36.04         & 94.82     & 18.92 & 37.25     	& 36.14\\ \midrule
MOM    			& 94.71     & 29.52	& 39.21   	& 40.89         & 94.82    	& 21.40 & 37.68    		& 37.26          \\
MOM ($\pm 1$)   & 94.71    & 38.20	& 40.99    	& 45.03         & 94.82    	& 29.14 & 39.26     & 40.90          \\
SANDY (Oracle)    & \textbf{96.47}     & \textbf{65.40} & \textbf{44.03}  & \textbf{56.48}         	& \textbf{96.42}     & \textbf{65.55} & \textbf{44.09}     & \textbf{56.62}         \\
SANDY (OCR)    & \textbf{96.47}     & 37.82 & 41.50  & 45.77         & \textbf{96.42}    & 37.78 & 41.49     & 45.81         \\ \bottomrule
\end{tabular}
\end{table*}

\begin{table*}[t]
\centering
\footnotesize
\caption{Results for chart-specific questions and answers. State-of-the-art VQA algorithms struggle with both.}
\vspace{-6pt}
\label{tbl:ans-type}
\begin{tabular}{@{}lrrr|rrr@{}}
\toprule
           & \multicolumn{3}{c|}{\textbf{Test-Familiar}}                                                                                                                                                                                                                              & \multicolumn{3}{c}{\textbf{Test-Novel}}                                                                                                                                                                                                                                \\ \midrule
           & \multicolumn{1}{c}{\multirow{2}{*}{\begin{tabular}[c]{@{}c@{}}Chart-specific\\ Questions\end{tabular}}} & \multicolumn{1}{c}{\multirow{2}{*}{\begin{tabular}[c]{@{}c@{}}Chart-specific\\ Answers\end{tabular}}} & \multicolumn{1}{c|}{\multirow{2}{*}{\textbf{Overall}}} & \multicolumn{1}{c}{\multirow{2}{*}{\begin{tabular}[c]{@{}c@{}}Chart-specific\\ Questions\end{tabular}}} & \multicolumn{1}{c}{\multirow{2}{*}{\begin{tabular}[c]{@{}c@{}}Chart-specific\\ Answers\end{tabular}}} & \multicolumn{1}{c}{\multirow{2}{*}{\textbf{Overall}}} \\
           & \multicolumn{1}{c}{}                                                                                    & \multicolumn{1}{c}{}                                                                                  & \multicolumn{1}{c|}{}                         & \multicolumn{1}{c}{}                                                                                    & \multicolumn{1}{c}{}                                                                                  & \multicolumn{1}{c}{}                         \\ \midrule
YES        & 17.71                                                                                                   & 0.00                                                                                                  & 11.78                                         & 17.58                                                                                                   & 0.00                                                                                                  & 11.77                                        \\
IMG        & 17.61                                                                                                   & 0.00                                                                                                  & 14.83                                         & 17.88                                                                                                   & 0.00                                                                                                  & 14.90                                        \\
QUES       & 23.17                                                                                                   & 0.10                                                                                                  & 21.06                                         & 22.97                                                                                                   & 0.00                                                                                                  & 21.00                                        \\
IMG+QUES   & 25.52                                                                                                   & 0.09                                                                                                  & 32.01                                         & 25.49                                                                                                   & 0.00                                                                                                  & 32.01                                        \\
SAN-VQA    &  26.54                                                                                           &   0.10                                                                                               &36.04                                       &26.32                                                                                              &0.00                                                                                                  &36.14                                          \\ \midrule
MOM        & 26.54                                                                                                    & 12.78                                                                                                  &40.89                                 & 26.32                                                                                              &2.93                                                                                                  & 37.26                                          \\
MOM ($\pm 1$)   & 26.54                                                                                                    &23.62                                                                                                 & 45.03                                         & 26.32                                                                                                     & 12.47                                                                                                  &40.90                                         \\
SANDY (Oracle)      & \textbf{27.80}                                                                                                  & \textbf{52.55}                                                                                                & \textbf{56.48}                                        & \textbf{27.77}                                                                                                   & \textbf{52.70}                                                                                               & \textbf{56.62}                                        \\
SANDY (OCR)    & 26.60     & 25.19 & 45.77  & 26.43         & 25.12    & 45.81         \\ \bottomrule
\end{tabular}
\end{table*}

\section{Experiments}

\begin{figure*}[t]
\centering
    \captionsetup[subfigure]{labelformat=empty}    
        \begin{subfigure}[t]{0.32\textwidth}
		\includegraphics[width=\textwidth]{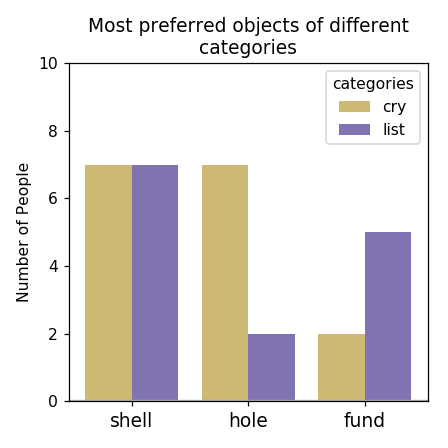}
        \caption{\footnotesize	    \textbf{Q}: How many objects are preferred by less than 7 people in at least one category?\\
        \textcolor{red}{SAN}: two \cmark ~ \textcolor{blue}{MOM}: two \cmark ~ \textcolor{magenta}{SANDY}: two\cmark\\ 
        \textbf{Q}: What category does the medium purple color represent?\\
        \textcolor{red}{SAN}: closet \xmark ~ \textcolor{blue}{MOM}: lisit \xmark ~ \textcolor{magenta}{SANDY}: list \cmark\\
      }
    \end{subfigure}
    \hfill
    \begin{subfigure}[t]{0.33\textwidth}
	 	\includegraphics[width=\textwidth]{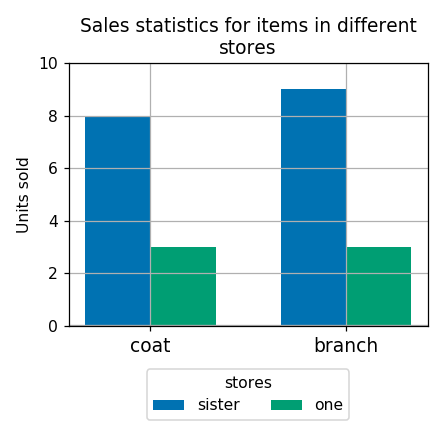}
\caption{ \footnotesize \textbf{Q}: Which item sold the most number of units summed across all the stores?\\
        \textcolor{red}{SAN}: closet\xmark~  \textcolor{blue}{MOM}: branch\cmark~ \textcolor{magenta}{SANDY}: branch\cmark\\
	    \textbf{Q}: How many units of the item branch were sold in the store sister?\\
        \textcolor{red}{SAN}: 9 \cmark  ~ \textcolor{blue}{MOM}: 9 \cmark  ~ \textcolor{magenta}{SANDY}: 9 \cmark\\  }
	\end{subfigure}
    \hfill
        \begin{subfigure}[t]{0.32\textwidth}
	 	\includegraphics[width=\textwidth]{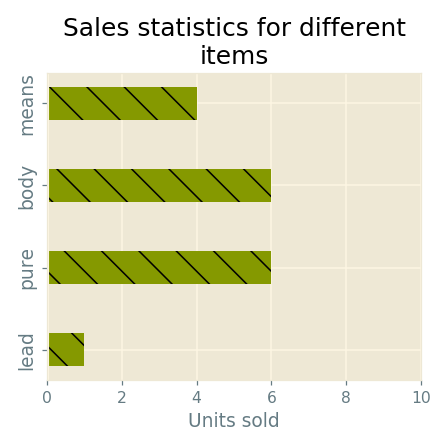}
\caption{\footnotesize \textbf{Q}: Are the values in the chart presented in a percentage scale?\\
        \textcolor{red}{SAN}: no \cmark~  \textcolor{blue}{MOM}: no \cmark~\textcolor{magenta}{SANDY}: no \cmark\\
	    \textbf{Q}: How many units of items lead and pure were sold?\\
        \textcolor{red}{SAN}: 8 \xmark ~ \textcolor{blue}{MOM}: 8 \xmark ~ \textcolor{magenta}{SANDY}: 7 \cmark\\       }
	\end{subfigure}
    \vspace{-20pt}
    \caption{\small Example results for different models on DVQA. Outputs of oracle version of SANDY model are shown. SAN completely fails to predict chart-specific answers whereas MOM model often makes small OCR errors (left). Both MOM and SAN are also incapable of properly encoding chart-specific labels in questions (right).}
    \label{fig:example-output}
    \end{figure*}

In this section, we describe the experimental results for models trained and tested on the DVQA dataset. DVQA's extensive annotations are used to analyze the performance of each model on different question- and answer-types to reveal their respective strengths and  weaknesses. In our experiments, we study the performance of algorithms on both familiar and novel chart labels, which are contained in two distinct test splits, Test-Familiar and Test-Novel. Every bar chart in Test-Familiar contains only labels seen during training. All of the models using the LSTM-encoder have entries in their word dictionaries for these familiar words, and all answers have been seen in the training set. The labels for the charts in Test-Novel are only  seen in the test set, and no system has them in the dictionaries they use to encode words or to generate answers.

To measure performance, an algorithm gets a question correct only if it generates a string that is identical to the ground truth. To better assess MOM, we also measure its performance using edit distance, which is denoted MOM ($\pm 1$). This model is allowed to get a question correct as long as the answer it generates is within one edit distance or less compared to the correct answer.

\subsection{General Observations}

Overall performance of each method broken down based on question-type are shown in 
Table~\ref{tbl:ques-type} and some qualitative examples are shown in Fig~\ref{fig:example-output}. Across all question-types, NO, IMG, and QUES  are the first, second, and third worst performing, respectively. Overall, SANDY performs best on both Test-Familiar and Test-Novel with SANDY-real following closely behind.

For structure questions, there is little difference across models for Test-Familiar and Test-Novel, which is expected because these questions ask about the general visual organization of a chart and do not require label reading. Performance increases greatly for IMG+QUES compared to either IMG or QUES, indicating structure questions demand combining image and question features.

For data retrieval and reasoning questions, SANDY and MOM both outperformed all baseline models. Both SANDY and SANDY-real outperformed MOM, and this gap was greater for Test-Novel.

\subsection{Chart-specific Words in Questions and Answers}
\new{
Many DVQA questions have chart-specific answers, \eg, `Which algorithm has the highest accuracy?' needs to be answered with the label of the bar with the highest value. These chart-specific answers are different than the generic answers that are shared across many bar charts, \eg, `Does the chart contain stacked bars?'. Similarly, some DVQA questions refer to elements that are specific to a given chart, \eg, `What is the accuracy of the algorithm \textbf{A}?'. To accurately answer these questions, an algorithm must be able to interpret the text-label \textbf{A} in the context of the given bar chart. Table~\ref{tbl:ans-type} shows the accuracy of the algorithms for questions that have chart-specific labels in them (chart-specific questions) and questions whose answers contain chart-specific labels (chart-specific answers). As shown, whenever chart-specific labels appear in the answer, both IMG+QUES and SAN-VQA fail abysmally. While this is expected for Test-Novel, they perform no better on Test-Familiar. Likewise, all of the models except SANDY also face difficulty for questions with chart-specific labels. Overall, they fail to meaningfully outperform the QUES baseline. We believe that the small gain in accuracy by IMG+QUES and SAN-VQA over QUES is only because the image information, such as the type of scale used (linear, percentage, or logarithmic), enables these methods to guess answers with higher precision.

In chart-specific answers, SANDY showed highest accuracy. Moreover, its performance for Test-Novel is similar to that for Test-Familiar. In comparision, while MOM outperforms the baselines, its accuracy on Test-Novel is much lower than its accuracy on Test-Familiar. This could be because MOM's string generation system is unable to produce accurate results with novel words. Supporting this, MOM often makes small string generation
errors, as shown by the improved performance of MOM~($\pm 1$), which is evaluated using edit distance. MOM's output is also dependent on the precise prediction of the bounding box containing the answer which could further affect the final accuracy. MOM's localization performance is explored in more detail in the supplemental materials.

In addition to SANDY's ability to predict chart-specific answer tokens, it can also be used to properly tokenize the chart-specific words in questions. An LSTM based question encoder using a fixed vocabulary will not be able to encode the questions properly, especially when encoding questions with unknown words in Test-Novel. For questions with chart-specific labels on them, SANDY shows improvement in properly encoding the questions with the chart-specific labels compared to baselines. However, the improvement in performance is not as drastic as seen for chart-specific answers. This may be due to the fact that many of the chart-specific questions include precise measurement \eg `How many people prefer object \textbf{O}?' which could be beyond the capacity of the SAN architecture.}

\section{Discussion}

In this paper we presented the DVQA dataset and explored models for DVQA. Our experiments show that VQA algorithms are only capable of answering simple structure questions. They perform much more poorly on data retrieval and reasoning questions, whereas our approaches, SANDY and MOM, are able to better answer these questions. Moreover, SANDY and MOM can both produce answers that are novel to the test set, which is impossible for traditional VQA algorithms. Finally, SANDY can also encode questions with novel words from the bar chart.

We studied SANDY's performance using a real OCR and a perfect oracle OCR system. While the performance dropped when real OCR was used, it still surpassed other algorithms across all categories. Despite its success, the proposed dynamic encoding used in SANDY is simple and offers a lot of room for expansion. Currently, the dynamic encoding is inferred based on the position of previously detected words. Any error in the OCR system in detecting a single word will propagate throughout the chain and render the encoding for the whole image useless. While this is not a problem for a perfect OCR, removing the cascaded reliance on correctness of each OCR results can help improve performance for an imperfect real-world OCR system.

Recently, multiple compositional models for VQA, such as neural module networks (NMN)~\cite{AndreasRDK15,andreas2016learning,hu2017learning}, have been developed. These recursive neural network systems consist of stacked sub-networks that are executed to answer questions, and they work well on the compositional reasoning questions in CLEVR~\cite{johnson2016clevr}. However, current NMN formulations are unable to produce chart-specific answers, so they cannot be used for DVQA without suitable modifications. 

SANDY and MOM are both built on top of SAN, and they try to solve chart-specific answer generation in two distinct ways that are agnostic to SAN's actual architecture. SANDY uses DEM and OCR to encode an image's text, whereas MOM attempts to predict the location of the text it needs to generate an answer. As VQA systems continue to evolve, upgrading SAN with an better VQA algorithm could improve the performance of our systems.

Our dataset currently only contains bar charts. We are developing a follow-up version that will contain pie charts, plots, and other visualizations in addition to bar-charts. Since neither MOM nor SANDY are designed specifically for bar-charts, they can operate on these alternative diagrams with only minor modifications.

\new{
We conducted an additional study to assess how well these models work on real bar charts. We manually annotated over 500 structure understanding questions for real bar charts scraped from the Internet. Without any fine-tuning, all of the SAN-based models achieved about 59\% accuracy on these questions, a 15\% absolute improvement over the image blind (QUES) baseline. This shows a positive transfer from synthetic to real-world bar charts. Training on entirely real charts would be ideal, but even then there would likely be a benefit to using synthetic datasets as a form of data augmentation~\cite{kafle2017data}.}

\section{Conclusion}
Here, we described DVQA, a dataset for understanding bar charts. We demonstrated that VQA algorithms are incapable of answering simple  DVQA questions. We proposed two DVQA algorithms that can handle chart-specific words in questions and answers. Solving DVQA will enable systems that can be utilized to intelligently query massive repositories of human-generated data, which would be an enormous aid to scientists and businesses. We hope that the DVQA dataset, which will be made publicly available, will promote the study of issues that are generally ignored with VQA with natural images, \eg, out-of-vocabulary words and dynamic question encoding. We also hope that DVQA will serve as an important proxy task for studying visual attention, memory, and reasoning capabilities.

{\small
\bibliographystyle{ieee}
\bibliography{egbib}
}

\begin{appendices}

\section{Additional details about the dataset}\label{sec:dataset}
In this section, we present additional details on the DVQA dataset statistics and how it was generated.
\begin{figure*}[t]
\centering
\footnotesize
    \captionsetup[subfigure]{labelformat=empty}    
        \begin{subfigure}[t]{0.22\textwidth}
		\includegraphics[width=\textwidth,]{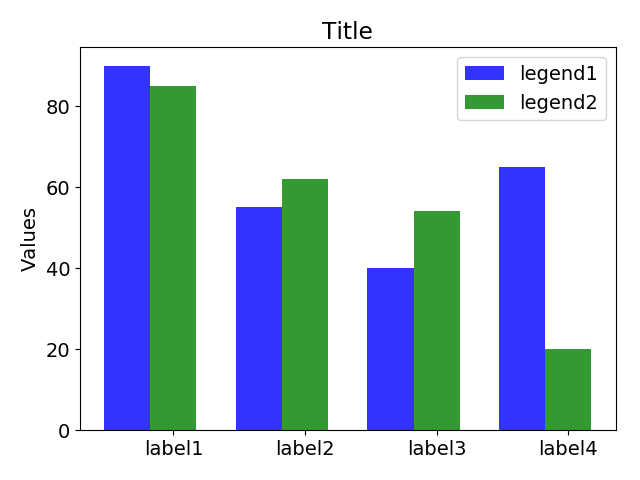}
        \caption{What is the value of \textit{label1} in \textit{legend2}?}
    \end{subfigure}
    \hfill
    \begin{subfigure}[t]{0.22\textwidth}
	 	\includegraphics[width=\textwidth ]{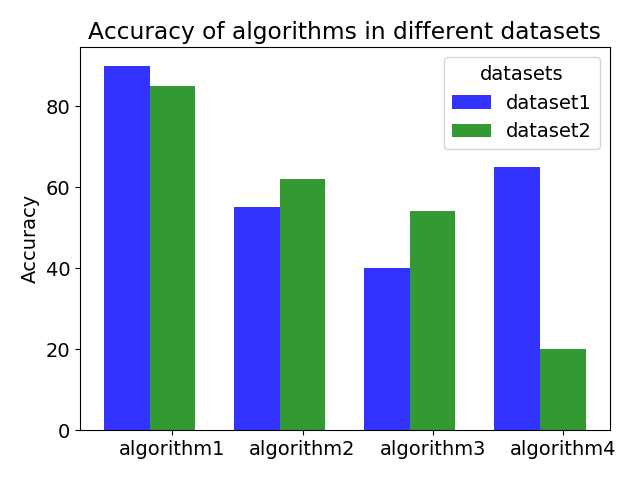}
         \caption{What is the accuracy of the algorithm \textit{algorithm1} in the dataset \textit{dataset2}?}

	\end{subfigure}
    \hfill
        \begin{subfigure}[t]{0.22\textwidth}
	 	\includegraphics[width=\textwidth ]{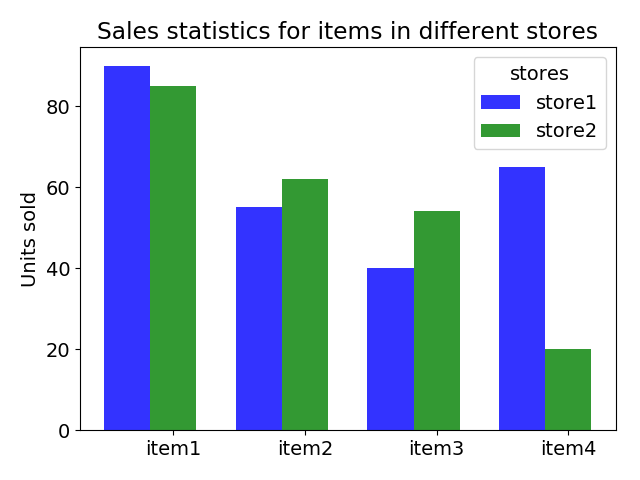}        
        \caption{How many units of the item \textit{item1} was sold in the store \textit{store2}?}

	\end{subfigure}
    \hfill
    \begin{subfigure}[t]{0.22\textwidth}
		\includegraphics[width=\textwidth]{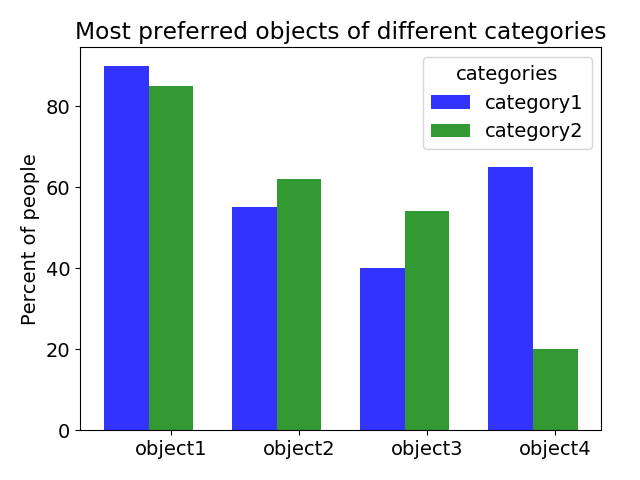}
        \caption{What percentage of people prefer the object \textit{object1} in the category \textit{category2}?}

    \end{subfigure}
    \caption{An example showing that different question can be created by using different title and labels in the same chart.}
    \label{fig:example-variations}
    \end{figure*}
\subsection{Data statistics}

Table~\ref{tbl:stat-qtypes} extends Table 1 of the main paper on the distribution of questions in the DVQA dataset.

\begin{table*}[t]
\centering
\caption{Statistics on different splits of dataset based on different question types.}
\label{tbl:stat-qtypes}
\begin{tabular}{@{}llrrr@{}}
\toprule
                           &               & \textbf{\begin{tabular}[c]{@{}l@{}}Total\\ Questions\end{tabular}} & \textbf{\begin{tabular}[c]{@{}l@{}}Unique\\ Answers\end{tabular}} & \textbf{\begin{tabular}[c]{@{}l@{}}Top-2 Answers\\ (in percentage) \end{tabular}} \\ \midrule
\multirow{3}{*}{Structure} & Train         & 313,842                                                             & 10     & no: 40.71, yes: 40.71                                                    \\
                           & Test-Familiar & 78,278                                                             & 10       & no: 41.14, yes: 41.14                                                     \\
                           & Test-Novel    & 78,988                                                             & 10     & no: 41.00, yes: 41.00	                                                \\ \midrule
\multirow{3}{*}{Data}      & Train         & 742,896                                                             & 1038    & no: 7.55, yes: 7.55 		                                                     \\
                           & Test-Familiar & 185,356                                                          & 1038      & no: 7.44, yes: 7.44 		                                               \\
                           & Test-Novel    & 185,452                                                             & 538      & no: 7.51, yes: 7.51		                                             \\ \midrule
\multirow{3}{*}{Reasoning} & Train         & 1,076,391                                                            & 1076        &yes: 8.29, no: 8.26                                                     \\
                           & Test-Familiar & 268,795                                                            & 1075       & no: 8.31, yes: 8.27                                                   \\
                           & Test-Novel    & 268,788                                                             & 577          &no: 8.28, yes: 8.22                                        \\ \midrule
\multirow{3}{*}{\textbf{Overall}} & Train         & 2,325,316                                                             & 1076        & yes: 11.74, no: 11.73                                                    \\
                           & Test-Familiar & 580,557                                                             & 1075         & yes: 11.77, no: 11.75                                               \\ & Test-Novel    & 581,321   & 577         & no: 11.80, yes: 11.77 

\\ \bottomrule
\end{tabular}
\end{table*}

\subsection{Variations in question templates}
The meaning of different entities in a chart is determined by its title and labels. This allows us to introduce variations in the questions by changing the title of the chart. For example, for a generic title `Title' and a generic label `Values', the base-question is: `What is the value of \textbf{L}?'. Depending on the title of the chart, the same question can take following forms:
\begin{enumerate}
\item Title: Accuracy of different algorithms, Label: Accuracy $\Rightarrow$ What is the accuracy of the algorithm \textbf{A}?
\item Title: Most preferred objects, Label: Percentage of people $\Rightarrow$  What percentage of people prefer object \textbf{O}?
\item Title: Sales statistics of different items, Label: Units sold $\Rightarrow$  How many units of the item \textbf{I} were sold?
\end{enumerate}

Figure ~\ref{fig:example-variations} provides an example on how questions can be varied for the same chart by using a different title and different labels.
\subsection{Data and visualization generation}

In this section, we provide additional details on the heuristics and methods used for generating question-answer pairs.

We aim to design the DVQA dataset such that commonly found visual and data patterns are also more commonly encountered in the DVQA dataset. To achieve this, we downloaded a small sample of bar-charts from Google image search and \textit{loosely} based the distribution of our DVQA dataset on the distribution of downloaded charts. However, some types of chart elements such as logarithmic axes, negative values, etc.\ that do not occur frequently in the wild are still very important to be studied. To incorporate these in our dataset, we applied such chart elements to a small proportion of the overall dataset. However, we made sure that each of the possible variations was encountered at least 1000 times in the training set.

\subsubsection{Distribution of visual styles} 
To incorporate charts with several appearances and styles in our DVQA dataset, we introduced different types of variations in the charts. Some of them as listed below:

\begin{enumerate}
\item Variability in the number of bars and/or groups of bars.
\item Single-column vs.\ multi-column grouped charts.
\item Grouped bars vs.\ stacked bars. Stacked bars are further divided into two types: 1) Additive stacking,  where bars represent individual values, and 2) Fractional stacking, where each bar represents a fraction of the whole.
\item Presence or absence of grid-lines.
\item Hatching and other types of textures.
\item Text label orientation.
\item A variety of colors, including monochrome styles.
\item Legends placed in a variety of common positions, including legends that are separate from the chart.
\item Bar width and spacing.
\item Varying titles, labels, and legend entries.
\item Vertical vs.\ horizontal bar orientation.
\end{enumerate}

In the wild, some styles are more common than others. To reflect this in our DVQA dataset, less common styles, \eg\ hatched bars, are applied to only a small subset of charts. However, every style-choice appears at least a 1000 times in the training set.
In overall, 70\% of the charts have vertical bars and the remaining charts have horizontal bars. Among multi-column bar-charts, 20\% of the linear and normalized percentage bar-charts are presented as stacked bar-charts and the rest are presented as group bar-charts. In legends we have used two styles that are commonly found in the wild: 1) legend below the chart, and 2) legend to the right of the chart. In 40\% of the multi-column charts, legends are positioned outside the bounds of the main chart.  Finally, 20\% of the charts are hatch-filled with a randomly selected pattern out of six commonly used patterns (stripes, dots, circles, cross-hatch, stars, and grid).

\subsubsection{Distribution of data-types}
Our DVQA dataset contains three major types of data scales.

\begin{itemize}[noitemsep,nolistsep]
\item \textbf{Linear data.} Bar values are chosen from 1 -- 10, in an increment of 1. When bars are not stacked, the axis is clipped at 10. When bars are stacked, the maximum value of the axis is automatically set by the height of the tallest stack. For a small number of charts, values are randomly negated or allowed to have missing values (\ie value of zero which appears as a missing bar).

\item \textbf{Percentage data.} Bar values are randomly chosen from 10--100, in increments of 10. For a fraction of multi-column group bar charts with percentage data, we normalize the data in each group so that the values add up to 100, which is a common style. A small fraction of bars can also have missing or zero value.

\item \textbf{Exponential data.} Bar values are randomly chosen in the range of 1 - $10^{10}$. The axis is logarithmic.
\end{itemize}

The majority (70\%) of the data in the DVQA dataset is of the linear type (1--10). Among these, 10\% of the charts are allowed to have negative. Then, 25\% of the data contain percentage scales (10--100), among which half are normalized so that the percentages within each group add up to a 100\%. For 10\% of both linear and percentage data-type, bars are allowed to have missing  (zero) values. The remaining 5\% of the data is exponential in nature ranging from $10^{0}$ -- $10^{10}$.

\begin{figure}[t]
\footnotesize
    \captionsetup[subfigure]{labelformat=empty}    
        \begin{subfigure}[t]{0.23\textwidth}
		\includegraphics[width=\textwidth]{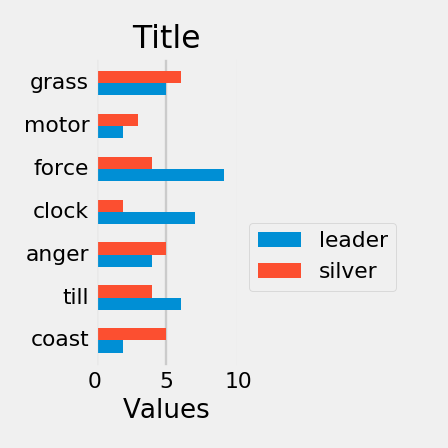}
    \end{subfigure}
    \hfill
    \begin{subfigure}[t]{0.23\textwidth}
	 	\includegraphics[width=\textwidth]{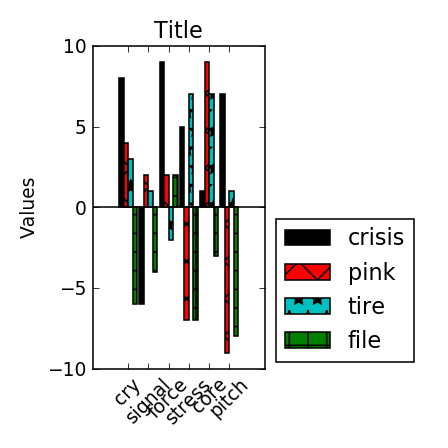}
	\end{subfigure}
    \caption{Examples of discarded visualizations due to the bar-chart being smaller than 50\% of the total image area.}
    \label{fig:discarded}
    \end{figure}

\subsubsection{Ensuring proper size and fit} Final chart images are drawn such that all of them have the same width and height of $448\times448$ pixels. This was done for the ease in processing and to ensure that the images do not need to undergo stretching or aspect ratio change when being processed using an existing CNN architecture. To attain this, we need to ensure that all the elements in the chart fit in the fixed image size. We have taken several steps to ensure a proper fit. By default, the label texts are drawn without rotation \ie horizontally. During this, if any of the texts overlap with each other, we rotate the text by either 45 or 90 degrees. Another issue is when the labels take up too much space leaving too little space for the actual bar-charts, which often makes them illegible. This is usually a problem with styles that contain large texts and/or charts where legend is presented on the side. To mitigate this, we discard the image if the chart-area is less than half of the entire image-area. Similarly, we also discard a chart if we cannot readjust the labels to fit without overlap despite rotating them. Fig.~\ref{fig:discarded} shows some examples of discarded charts due to poor fit.

\subsubsection{Naming colors} For generating diverse colors, we make use of many of the pre-defined styles that are available with the Matplotlib package and also modify it with several new color schemes. Matplotlib allows us to access the RBG face-color of each drawn bar and legend entries from which we can obtain the color of each of the element drawn in the image. However, to ask questions referring to the color of a bar or a legend entry, we need to be able to name it using natural language (\eg `What does the \texttt{red color} represent?). Moreover, simple names such as `blue' or `green' alone may not suffice to distinguish different colors in the chart. So, we employ the following heuristic to obtain a color name for a given RGB value.

\begin{enumerate}
\item Start with a dictionary of all 138 colors from the CSS3 X11 named colors. Each of the color is accompanied by its RGB value and its common name. The color names contain names such as darkgreen, skyblue, navy, lavender, chocolate, and other commonly used colors in addition to canonical color names such as `blue', `green', or `red'.

\item Convert all the colors to CIE standard L*a*b* color space which is designed to approximate human perception of the color space.
\item Measure color distance between the L*a*b* color of our chart-element and each of the color in the X11 color dictionary. For distance, we use the CIE 2000 delta E color difference measure which is designed to measure human perceptual differences between colors.
\item Choose the color from the X11 colors which has the lowest delta E value from the color of our chart-element.
\end{enumerate}

\section{Analysis of MOM's localization performance}\label{sec:analysis}

\begin{figure*}[t]
\centering
\footnotesize
    \captionsetup[subfigure]{labelformat=empty}    
        \begin{subfigure}[t]{0.32\textwidth}
		\includegraphics[width=\textwidth ,]{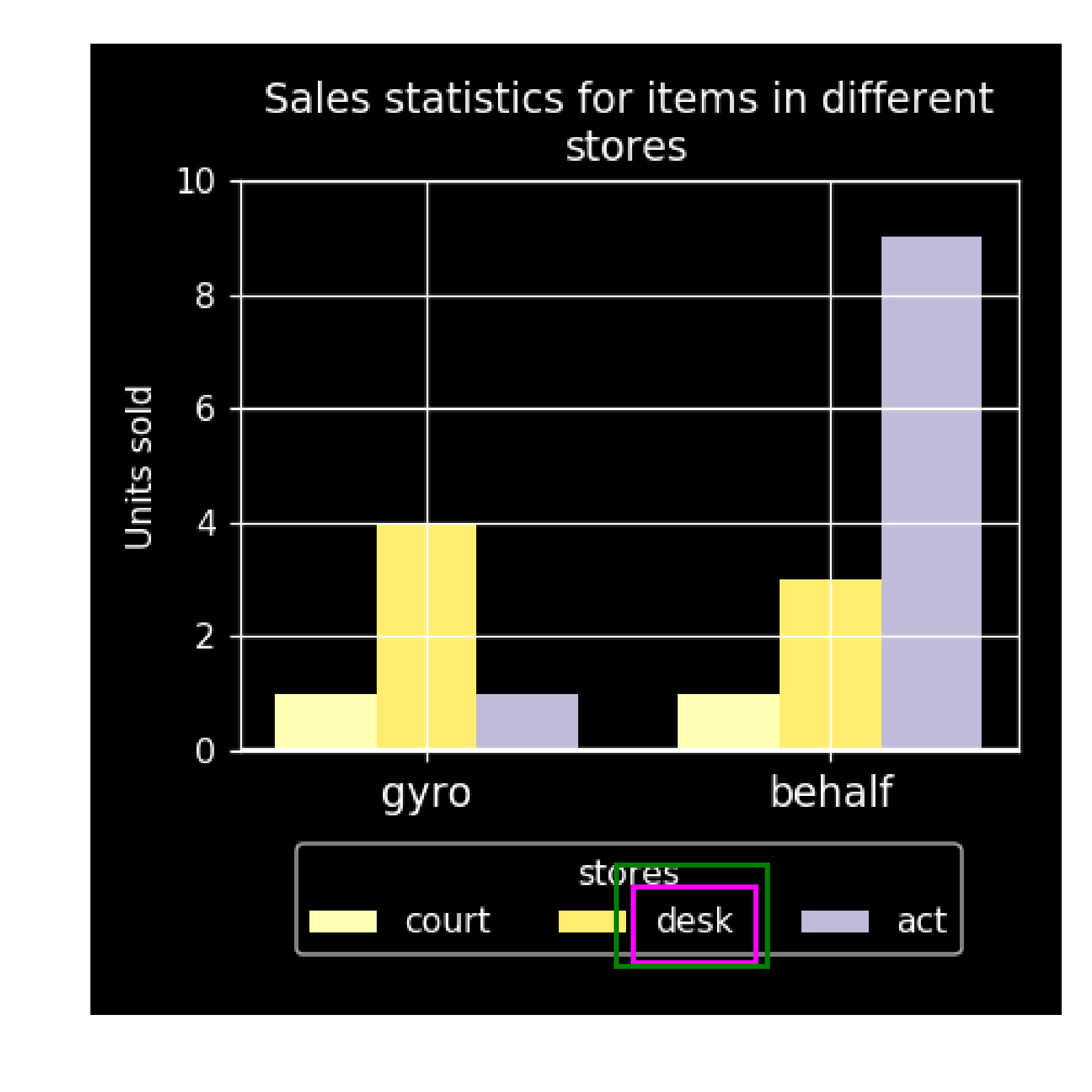}
    \end{subfigure}
    \hfill
    \begin{subfigure}[t]{0.32\textwidth}
	 	\includegraphics[width=\textwidth,]{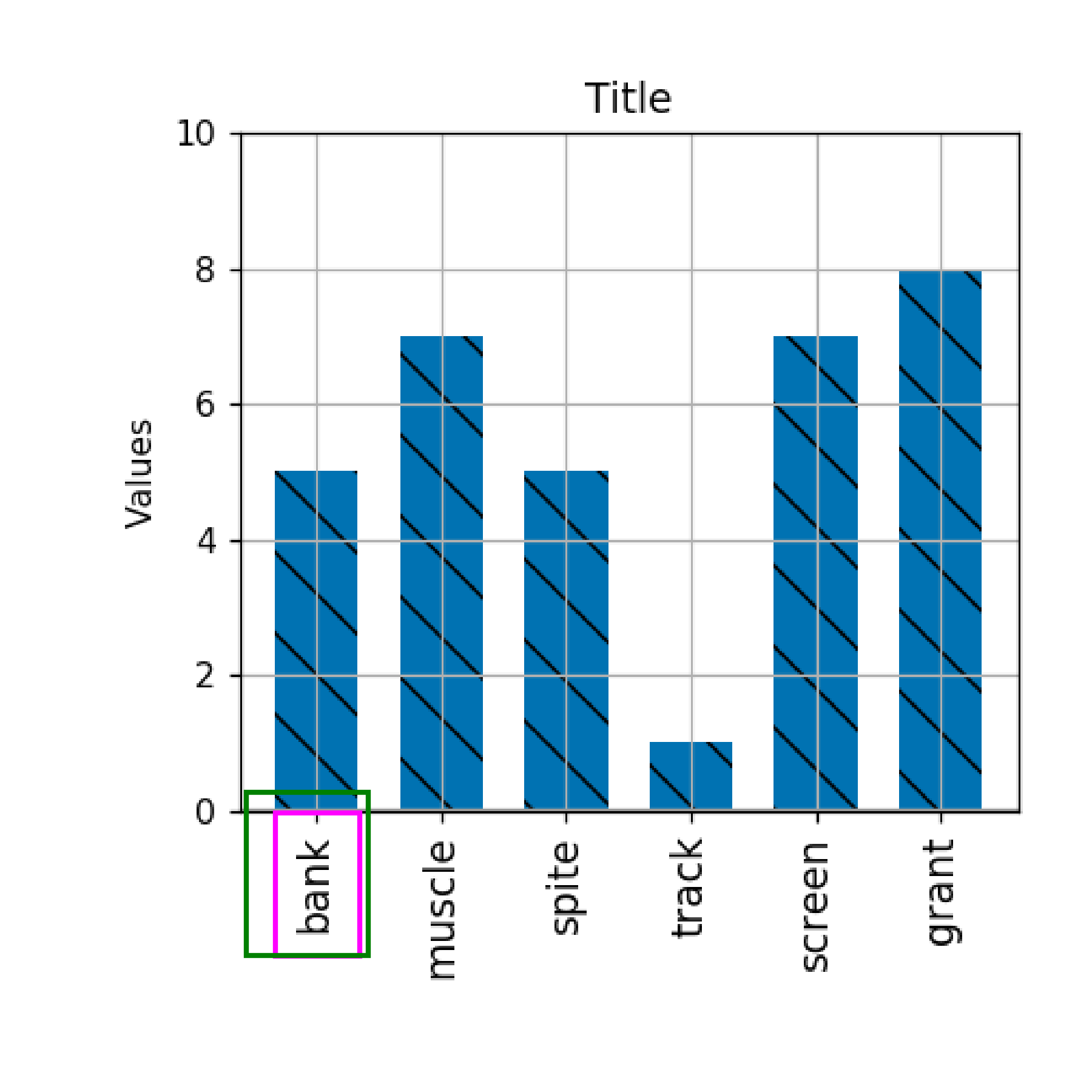}
	\end{subfigure}
    \hfill
        \begin{subfigure}[t]{0.32\textwidth}
	 	\includegraphics[width=\textwidth, ]{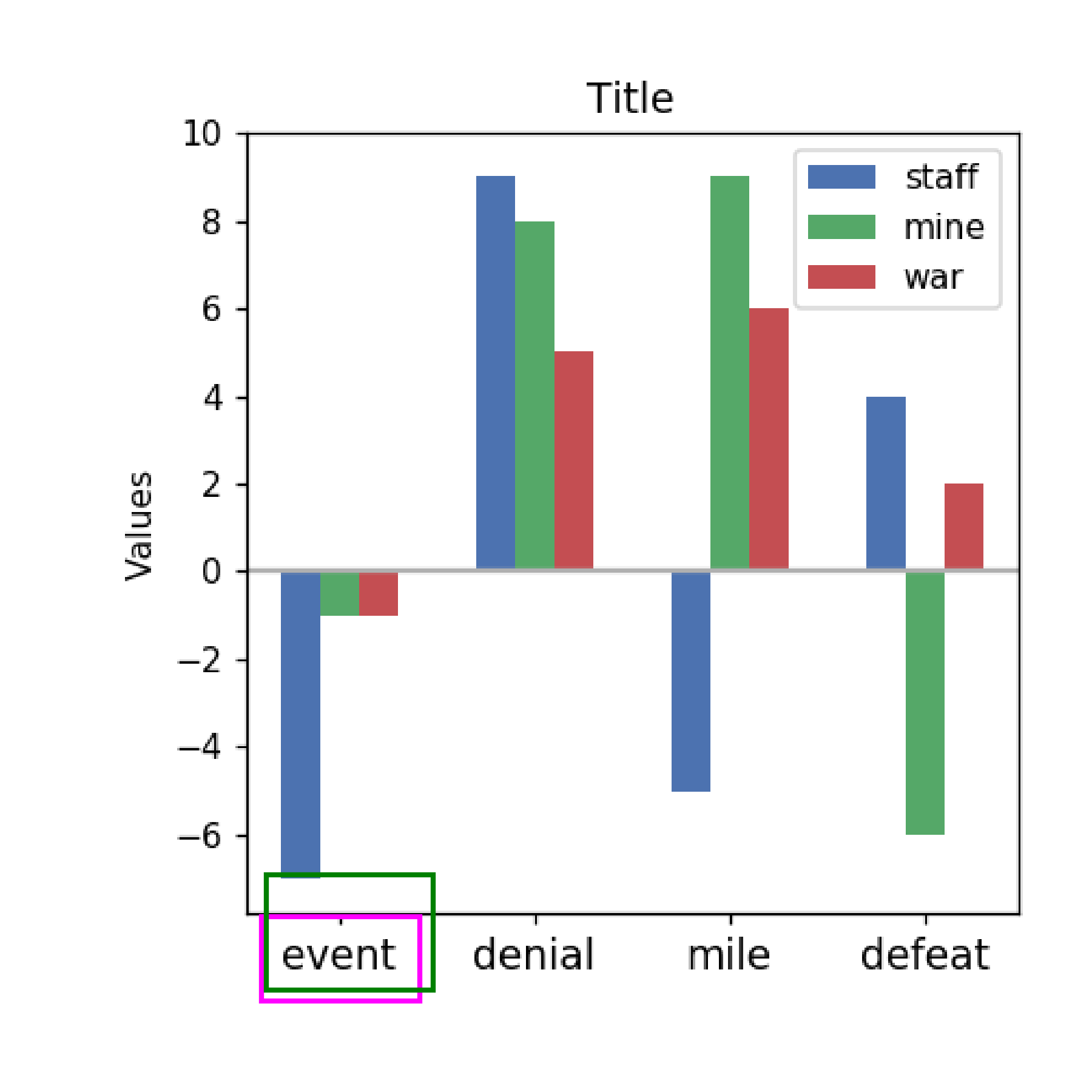}
	\end{subfigure}
    \caption{Some examples showing correctly predicted bounding boxes predicted by our MOM model. Magenta shows the ground truth and green shows the predicted bounding box.}
    \label{fig:example-correct}
    \end{figure*}

\begin{figure*}[t]
\centering
\footnotesize
    \captionsetup[subfigure]{labelformat=empty}    
        \begin{subfigure}[t]{0.32\textwidth}
		\includegraphics[width=\textwidth ,]{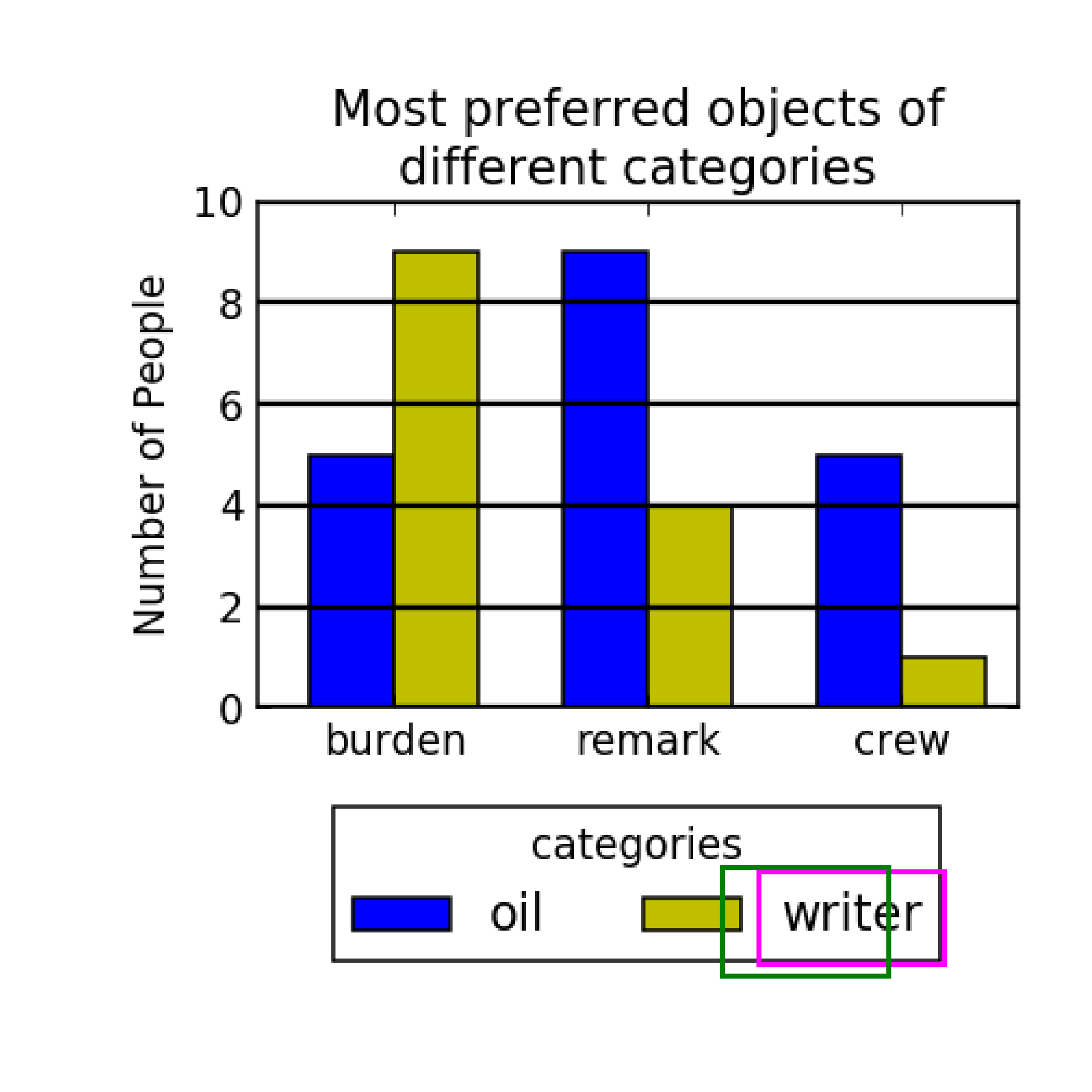}
    \end{subfigure}
    \hfill
    \begin{subfigure}[t]{0.32\textwidth}
	 	\includegraphics[width=\textwidth,]{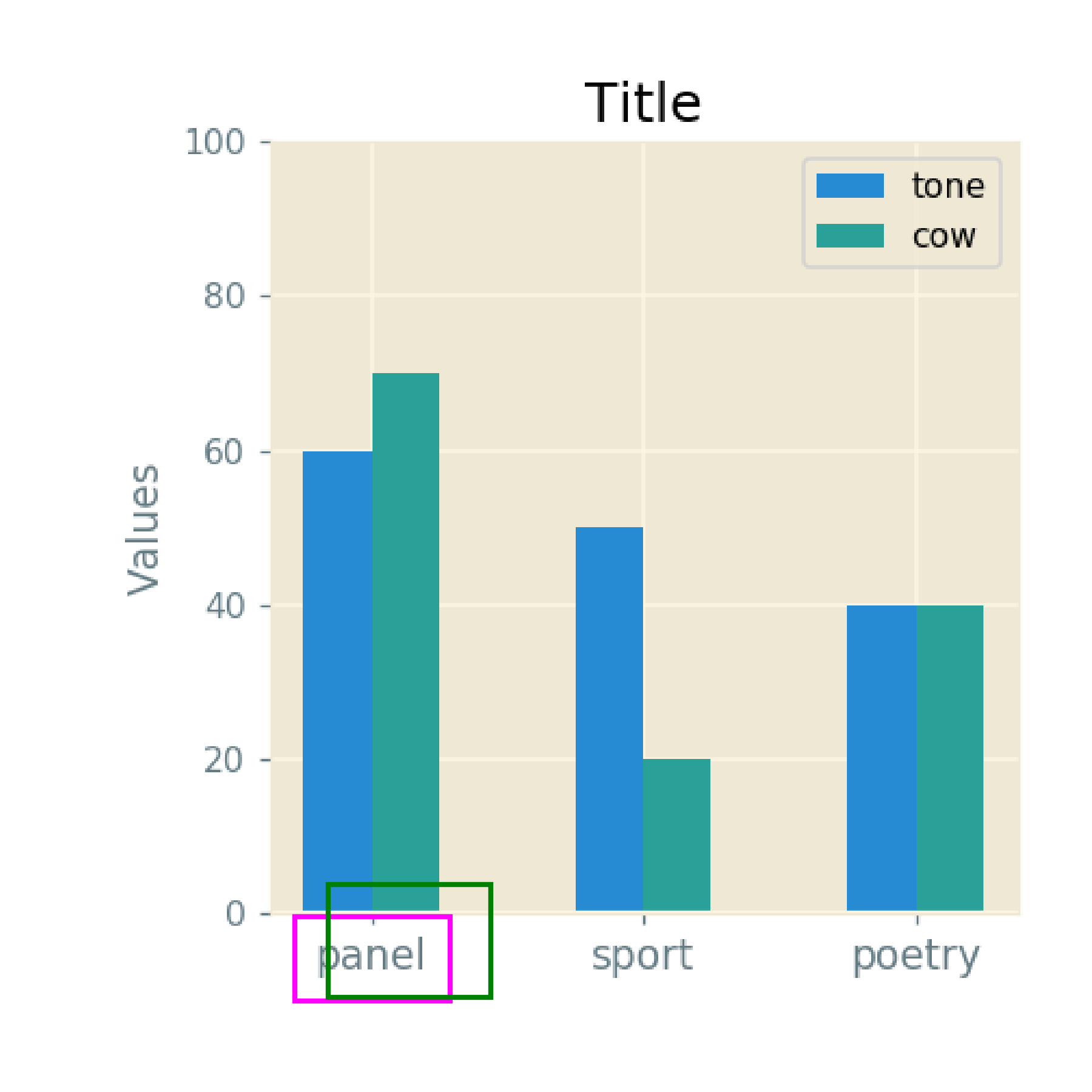}
	\end{subfigure}
    \hfill
        \begin{subfigure}[t]{0.32\textwidth}
	 	\includegraphics[width=\textwidth, ]{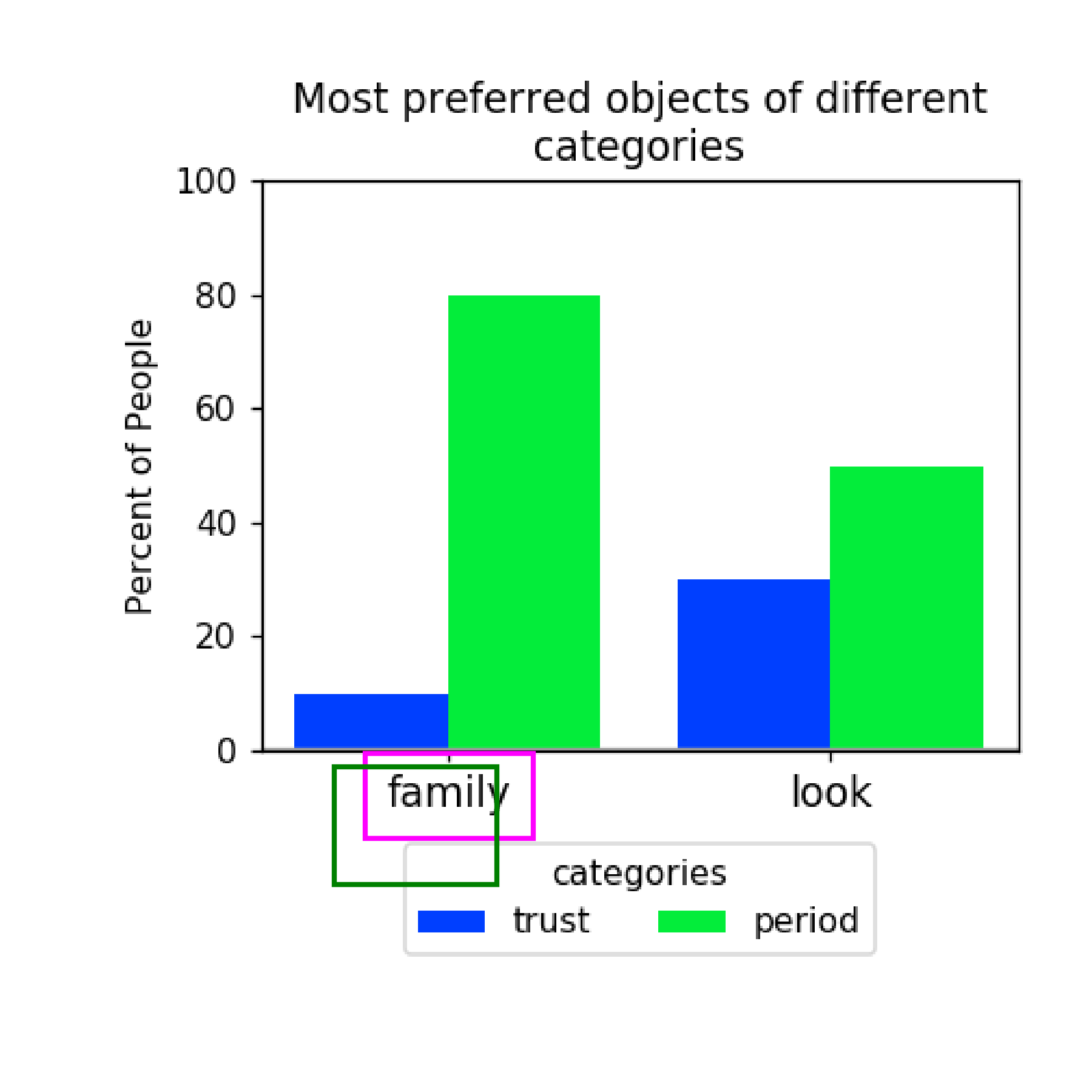}
	\end{subfigure}
    \caption{Some examples showing incorrectly predicted bounding boxes predicted by our MOM model. Often the prediction is off by only a few pixels, but the since the OCR requires total coverage, it results in an erroneous prediction. Magenta shows the ground truth and green shows the predicted bounding box.}
    \label{fig:example-slightly-wrong}
    \end{figure*}

\begin{table}[t]
\centering
\footnotesize
\caption{Localization performance of MOM in terms of IOU with the ground truth bounding box.}
\label{tbl:mom-localization-iou}
\begin{tabular}{@{}lll@{}}
\toprule
                 \textbf{\begin{tabular}[c]{@{}l@{}}IOU with\\ ground truth\end{tabular}}                & \textbf{\begin{tabular}[c]{@{}l@{}}Percentage\\ of boxes\end{tabular}} \\ \midrule
$\geq$ 0.2          &73.27     \\              
$\geq$ 0.4          &56.89     \\              
$\geq$ 0.5          &46.06     \\              
$\geq$ 0.6          &32.49     \\              
$\geq$ 0.7          &18.80     \\              
$\geq$ 0.8          &~6.93     \\              
$\geq$ 0.9          &~0.66     \\              
$\geq$ 1.0          &~0.00     \\                                                                       
\end{tabular}
\end{table}

\begin{table}[t]
\centering
\footnotesize
\caption{Localization performance of MOM in terms of the distance between the center of the predicted and ground truth bounding box.}
\label{tbl:mom-localization-dist}
\begin{tabular}{@{}lll@{}}
\toprule
                 \textbf{\begin{tabular}[c]{@{}l@{}}Distance from\\ the ground truth\end{tabular}}                & \textbf{\begin{tabular}[c]{@{}l@{}}Percentage\\ of boxes\end{tabular}} \\ \midrule
$\leq$ 1 pixels         &~0.14    \\              
$\leq$ 8 pixels          &~8.48     \\              
$\leq$ 16 pixels          &25.77     \\              
$\leq$ 32 pixels          &52.89     \\              
$\leq$ 64 pixels          &74.21     \\              
                                                                  
\end{tabular}
\end{table}

In the main paper, we observed that many predictions made by MOM were close to the ground truth but not exactly the same. This was also corroborated by taking into account the edit-distance between the predicted and ground truth answer strings. 

Here we study our hypothesis that this low accuracy is due to poor localization of the predicted bounding boxes. Fig.~\ref{fig:example-correct} shows some results from MOM for Test-Familiar split of the dataset in which the bounding boxes are accurately predicted. This shows that the bounding box prediction network works with texts of different orientations and positions. However, Fig.~\ref{fig:example-slightly-wrong} shows some examples where boxes do not `snap' neatly around the text area but are in the right vicinity. Since the OCR subnetwork in MOM operates only on the features extracted from the predicted bounding box, a poor bounding box would also translate to a poor prediction. To quantify this behavior we conduct two separate studies.

First, we measure the intersection over union (IOU) for predicted and ground truth bounding boxes. Table \ref{tbl:mom-localization-iou} shows the percentages of boxes that were accurately predicted for various threshold values of IOU.

Next, we measure what percentage of the predicted boxes are within a given distance from the ground truth boxes. The distance is measured as the Euclidean distance between the center x,y co-ordinates for predicted and ground-truth bounding boxes. Result presented in Table \ref{tbl:mom-localization-dist} shows that more than half of the predicted boxes are within 32 pixels from the ground truth boxes. Note here that the image dimension is 448$\times$448 pixels.

The above experiments show that while many of the predicted bounding boxes are `near' the ground truth boxes, they do not perfectly enclose the text. Therefore, if the predicted bounding boxes are localized better, which could be achieved with additional fine-tuning of the predicted bounding boxes, we can expect a considerable increase in MOM's accuracy on chart-specific answers.

\section{Additional examples}

\begin{figure*}[!h]
\centering
\footnotesize
    \captionsetup[subfigure]{labelformat=parens, labelsep=newline, justification=centering}    
        \begin{subfigure}[t]{0.32\textwidth}
		\includegraphics[width=\textwidth, ]{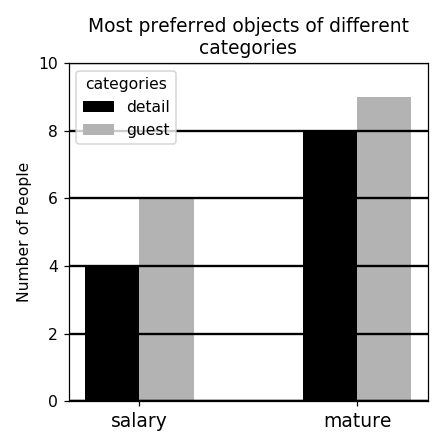}
        \caption{\textbf{Q}: What is the label of the second bar from the left in each group?\\
        \textcolor{red}{SAN}: closet \xmark ~  \textcolor{blue}{MOM}: guest \cmark ~ \textcolor{magenta}{SANDY}: guest \cmark\\
	    \textbf{Q}: Is each bar a single solid color without patterns?\\
        \textcolor{red}{SAN}: yes \cmark~ \textcolor{blue}{MOM}: yes \cmark~ \textcolor{magenta}{SANDY}: yes \cmark \\}
    \end{subfigure}
    \hfill
    \begin{subfigure}[t]{0.32\textwidth}
	 	\includegraphics[width=\textwidth, ]{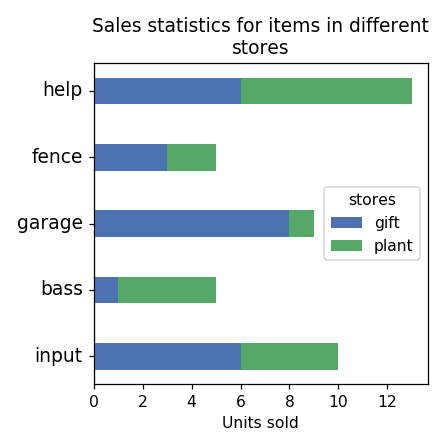}
\caption{\textbf{Q}: How many items sold less than 6 units in at least one store?\\
        \textcolor{red}{SAN}: four \cmark ~  \textcolor{blue}{MOM}: four \cmark~ \textcolor{magenta}{SANDY}: four\cmark \\
	    \textbf{Q}: Does the chart contain stacked bars?\\
        \textcolor{red}{SAN}: yes \cmark ~ \textcolor{blue}{MOM}: yes \cmark ~ \textcolor{magenta}{SANDY}: yes \cmark\\  }
	\end{subfigure}
    \hfill
        \begin{subfigure}[t]{0.32\textwidth}
	 	\includegraphics[width=\textwidth, ]{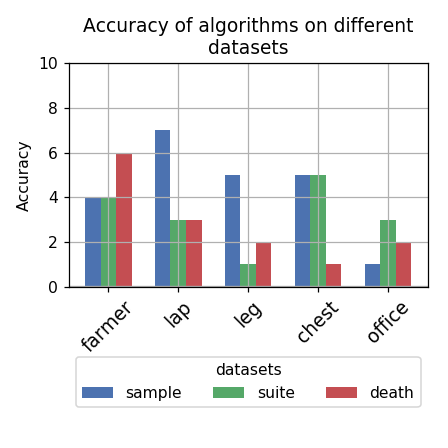}
\caption{\textbf{Q}: What is the highest accuracy reported in the whole chart?\\
        \textcolor{red}{SAN}: 7 \cmark ~  \textcolor{blue}{MOM}: 7 \cmark~ \textcolor{magenta}{SANDY}: 7 \cmark\\
	    \textbf{Q}: Is the accuracy of the algorithm leg in the dataset suite smaller than the accuracy of the algorithm chest in the dataset sample?\\
        \textcolor{red}{SAN}: no \xmark ~ \textcolor{blue}{MOM}: no \xmark ~ \textcolor{magenta}{SANDY}: yes \cmark \\   \label{fig:correct-c}    }
        \vspace{10px}
	\end{subfigure}
        \begin{subfigure}[t]{0.32\textwidth}
		\includegraphics[width=\textwidth, ]{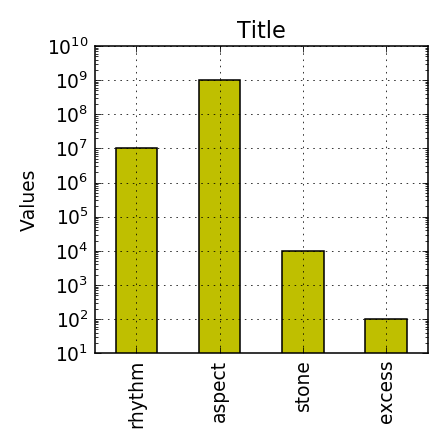}
        \caption{\textbf{Q}: Which bar has the largest value?\\
        \textcolor{red}{SAN}: closet \xmark ~\textcolor{blue}{MOM}: aspect \cmark ~  \textcolor{magenta}{SANDY}: aspect \cmark\\
	    \textbf{Q}: What is the value of the largest bar?\\
        \textcolor{red}{SAN}: $10^9$ \cmark ~ \textcolor{blue}{MOM}: $10^9$ \cmark ~ \textcolor{magenta}{SANDY}: $10^9$\cmark \\  }
    \end{subfigure}
    \hfill
    \begin{subfigure}[t]{0.32\textwidth}
	 	\includegraphics[width=\textwidth, ]{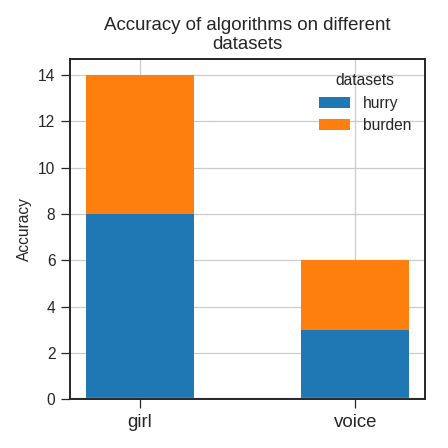}
\caption{\textbf{Q}: How many algorithms have accuracy lower than 3 in at least one dataset?\\
        \textcolor{red}{SAN}: zero \cmark ~  \textcolor{blue}{MOM}: zero \cmark ~ \textcolor{magenta}{SANDY}: zero \cmark\\
	    \textbf{Q}:  Which algorithm has highest accuracy for any dataset?\\
        \textcolor{red}{SAN}: closet \xmark  ~ \textcolor{blue}{MOM}: girl \cmark ~ \textcolor{magenta}{SANDY}: girl\cmark}
	\end{subfigure}
    \hfill
        \begin{subfigure}[t]{0.32\textwidth}
	 	\includegraphics[width=\textwidth, ]{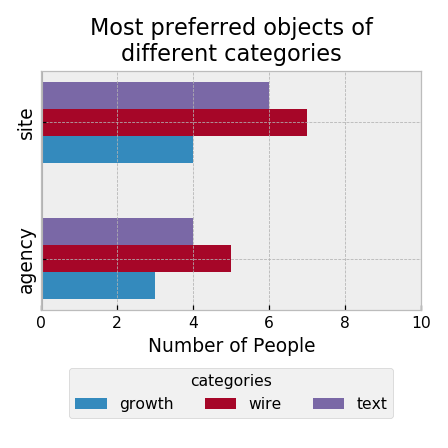}
\caption{\textbf{Q}: Which object is preferred by the most number of people summed across all the categories?\\
        \textcolor{red}{SAN}: closet \xmark~  \textcolor{blue}{MOM}: site \cmark~ \textcolor{magenta}{SANDY}: site\cmark\\
	    \textbf{Q}: Are the bars horizontal?\\
        \textcolor{red}{SAN}: yes \cmark ~ \textcolor{blue}{MOM}: yes \cmark ~ \textcolor{magenta}{SANDY}: yes \cmark \\       }
	\end{subfigure}
    \caption{Some example question-answer pair for different algorithms on the Test-Familiar split of the dataset.  The algorithms show success in variety of questions and visualizations. However, the SAN model is utterly incapable of predicting chart-specific answers.\label{fig:example-output-base}}
    \end{figure*}

\begin{figure*}[!h]
\centering
\footnotesize
    \captionsetup[subfigure]{labelformat=parens, labelsep=newline, justification=centering}    
        \begin{subfigure}[t]{0.32\textwidth}
		\includegraphics[width=\textwidth, ]{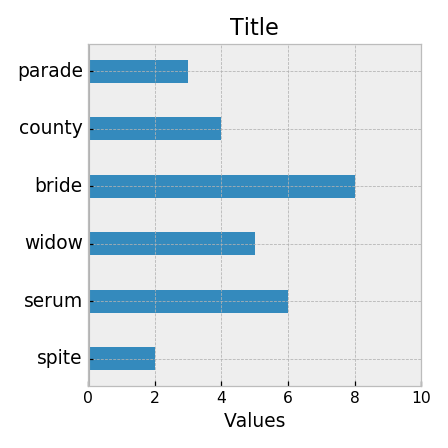}
        \caption{
	    \textbf{Q}: What is the label of the third bar from the bottom? \\
        \textcolor{red}{SAN}: closet \xmark ~ \textcolor{blue}{MOM}: whidkw \xmark ~ \textcolor{magenta}{SANDY}: widow \cmark\\  \label{subfig:a}}
    \end{subfigure}
    \hfill
    \begin{subfigure}[t]{0.32\textwidth}
	 	\includegraphics[width=\textwidth, ]{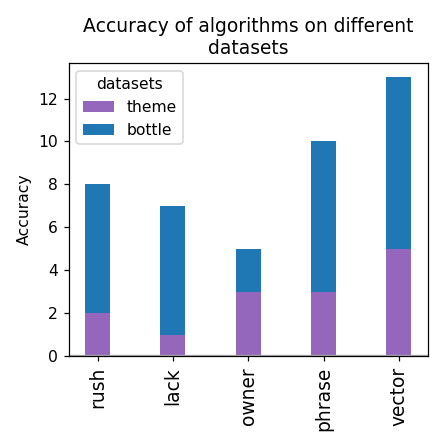}
        \caption{
	    \textbf{Q}: Which algorithm has the largest accuracy summed across all the datasets?\\
        \textcolor{red}{SAN}: closet \xmark  ~ \textcolor{blue}{MOM}: lack \xmark ~ \textcolor{magenta}{SANDY}: vector \cmark\\ \label{subfig:b} }
	\end{subfigure}
    \hfill
    \begin{subfigure}[t]{0.32\textwidth}
	 	\includegraphics[width=\textwidth, ]{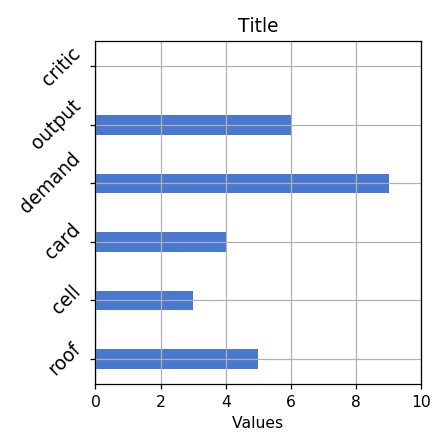}
\caption{\textbf{Q}: Is the value of output smaller than demand? \\
        \textcolor{red}{SAN}: no \xmark ~  \textcolor{blue}{MOM}: no \xmark ~ \textcolor{magenta}{SANDY}: yes \cmark\\
		\label{subfig:c} }
       \vspace{20px}
\end{subfigure}
        \begin{subfigure}[t]{0.32\textwidth}
		\includegraphics[width=\textwidth, ]{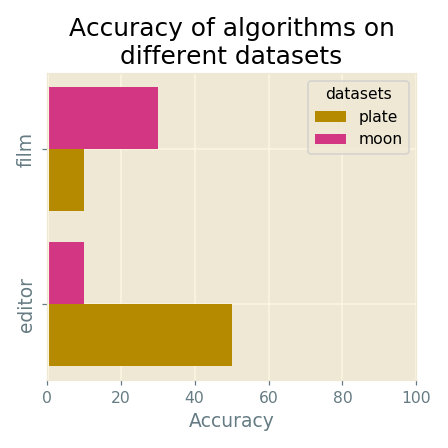}
        \caption{\textbf{Q}: Which algorithm has the smallest accuracy summed across all the datasets?\\
        \textcolor{red}{SAN}: closet \xmark ~  \textcolor{blue}{MOM}: fil \xmark ~ \textcolor{magenta}{SANDY}: editor \xmark\\
	    \textbf{Q}: What is the highest accuracy reported in the whole chart? \\
        \textcolor{red}{SAN}: 60 \xmark~ \textcolor{blue}{MOM}: 60 \xmark~ \textcolor{magenta}{SANDY}: 60 \xmark\\ \label{subfig:d} }
    \end{subfigure}
    \hfill
    \begin{subfigure}[t]{0.32\textwidth}
	 	\includegraphics[width=\textwidth, ]{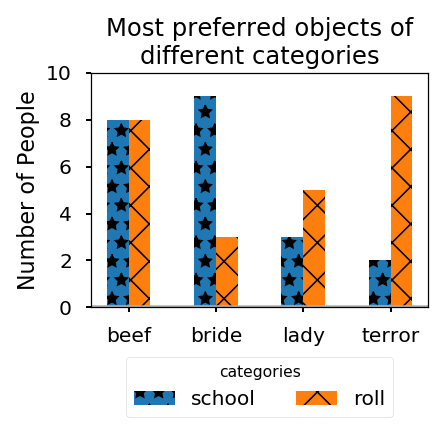}
\caption{\textbf{Q}: How many total people preferred the object terror across all the categories?\\
        \textcolor{red}{SAN}: 10 \xmark ~  \textcolor{blue}{MOM}: 10 \xmark ~ \textcolor{magenta}{SANDY}: 10 \xmark\\
	    \textbf{Q}: How many people prefer the object terror in the category roll?\\
        \textcolor{red}{SAN}: 1 \xmark ~ \textcolor{blue}{MOM}: 1 \xmark ~ \textcolor{magenta}{SANDY}: 9 \cmark \\ \label{subfig:e} }
	\end{subfigure}
    \hfill
        \begin{subfigure}[t]{0.32\textwidth}
	 	\includegraphics[width=\textwidth, ]{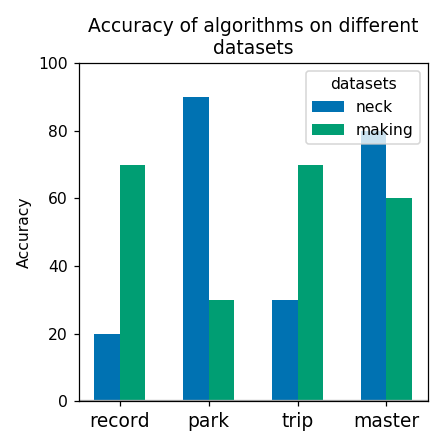}
\caption{\textbf{Q}: What is the highest accuracy reported in the whole chart?\\
        \textcolor{red}{SAN}: 90 \cmark ~  \textcolor{blue}{MOM}: 90 \cmark ~ \textcolor{magenta}{SANDY}: 80 \xmark \\
	    \textbf{Q}: Which algorithm has the smallest accuracy summed across all the datasets?\\
        \textcolor{red}{SAN}: closet \xmark ~  \textcolor{blue}{MOM}: record \cmark ~ \textcolor{magenta}{SANDY}: park \xmark \\      \label{subfig:f} }
	\end{subfigure}
    \caption{Some failure cases for different algorithms on the Test-Familiar split of the dataset. \label{fig:example-output-wrong}}
    \end{figure*}

In this section, we present additional examples to illustrate the performance of different algorithms for different types of questions. Fig.~\ref{fig:example-output-base} shows some example figures with question-answer results for different algorithms and Fig.~\ref{fig:example-output-wrong} shows some interesting failure cases.

As shown in Fig.~\ref{fig:example-output-base}, SAN-VQA, MOM, and SANDY all perform with high accuracy across different styles for structure understanding questions. This is unsurprising since all the models use the SAN architecture for answering these questions. However, despite the presence of answer-words in the training set (test-familiar split) SAN is incapable of answering questions with chart-specific answers; it always produces the same answer regardless of the question being asked. In comparison, MOM shows some success in decoding the chart-specific answers. However, as explained earlier in section \ref{sec:analysis}, the accuracy of MOM for chart-specific answers also depends on the accuracy of the bounding box prediction due to which its predictions were close but not exact for many questions. As discussed in section \ref{sec:analysis}, although the exact localization of the bounding box was poor, the majority of the predicted bounding boxes were in the vincinity of the ground truth bounding boxes. We believe with additional fine-tuning, \eg regressing for a more exact bounding box based on the features surrounding the initial prediction, could improve the model's performance significantly. Finally, SANDY shows a remarkable success in predicting the chart-specific answers. SANDY's dynamic dictionary converts the task of predicting the answer to predicting the position of the text in the image, making it easier to answer. Once the position is predicted, there are no additional sources of error for SANDY making it less error prone in general.

Similarly, both SAN and MOM are incapable of correctly parsing the questions with chart-specific labels in them. In comparision, SANDY can use the dynamic local dictionary to correctly parse the chart-specific labels showing an improved performance for these questions \eg Fig.~\ref{fig:correct-c}, \ref{subfig:c}, and \ref{subfig:e}.

In Fig.~\ref{fig:example-output-wrong}, we study some failure cases to better understand the nature of the errors made by current algorithms. One of the most commonly encountered errors for the algorithms that we tested is the error in predicting exact value of the data. Often, predicting these values involve extracting exact measurement and performing arithmetic operations across different values. The results show that the models are able to perform some measurement; the models predict values that are close to the correct answer, \eg predicting smaller values when the bars have smaller height (Fig.~\ref{subfig:d}) and predicting larger values when the bars are tall (Fig.~\ref{subfig:f}). In addition, the models are able to make predictions in the accurate data scale \eg For Fig.~\ref{subfig:d}, the prediction for the value is in percentage scale (0--100) and for Fig.~\ref{subfig:e}, the prediction is in linear scale (0--10).

The next class of the commonly encountered errors is the prediction of chart-specific answers. We have already established that the SAN-VQA model completely fails to answer questions with chart-specific answers, which is demonstrated in all the examples in Fig.~\ref{fig:example-output-base} and \ref{fig:example-output-wrong}. Our MOM model also makes errors for several examples as shown in Fig. \ref{fig:example-output-wrong}. The errors occur in decoding the OCR (Fig.~\ref{subfig:a}), predicting the right box (Fig.~\ref{subfig:f}) or both (Fig.~\ref{subfig:d}). While our SANDY model shows vastly increased accuracy for these answers, it can make occasional errors for these questions (Fig.~\ref{subfig:d}).
\vspace{100px}
\begin{table*}[!h]
\centering
\huge
\begin{tabular}{@{}lrrrrrr@{}}
\toprule
                     & \multicolumn{6}{c}{\textbf{Example question-answer pairs for different models}} \\ \bottomrule

\end{tabular}
\end{table*}

\begin{table*}[!h]
\centering
\huge
\begin{tabular}{@{}lrrrrrr@{}}
\toprule
                     & \multicolumn{6}{c}{\textbf{Some interesting failure cases}} \\ \bottomrule

\end{tabular}
\end{table*}

\end{appendices}

\end{document}